\let\texyear\year
\documentclass{ieeeaccess}
\let\ieeeaccessyear\year

\usepackage{cite}
\usepackage{amsmath,amssymb,amsfonts}
\usepackage{algorithmic}
\usepackage{graphicx}
\usepackage{textcomp}
\def\BibTeX{{\rm B\kern-.05em{\sc i\kern-.025em b}\kern-.08em
    T\kern-.1667em\lower.7ex\hbox{E}\kern-.125emX}}

\usepackage{blindtext}
\usepackage{multirow}

\let\year\texyear
    \usepackage{orcidlink}
\let\year\ieeeaccessyear

\let\svthefootnote\thefootnote
\newcommand\freefootnote[1]{
  \let\thefootnote\relax
  \footnotetext{#1}
  \let\thefootnote\svthefootnote
}

\begin{document}
\bstctlcite{IEEEexample:BSTcontrol}

\title{ClusterFusion: Leveraging Radar Spatial Features for Radar-Camera 3D Object Detection in Autonomous Vehicles}
\author{\uppercase{Irfan Tito Kurniawan} \uppercase{and} \uppercase{Bambang Riyanto Trilaksono}}
\address[]{School of Electrical Engineering and Informatics, Institut Teknologi Bandung, Bandung 40132, Indonesia}
\tfootnote{This work was supported in part by the Indonesian Ministry of Finance under the RISPRO LPDP (Pendanaan Riset Inovatif Produktif, Lembaga Pengelola Dana Pendidikan) Research Funding, Institut Teknologi Bandung under P2MI (Penelitian Pengabdian Masyarakat dan Inovasi) Research Program, and the Indonesian Ministry of Education, Culture, Research and Technology under the World Class University (WCU) Program managed by Institut Teknologi Bandung and under the World Class Research (WCR) Program.}

\markboth
{Kurniawan, I. T. and Trilaksono, B. R.: ClusterFusion: Leveraging Radar Spatial Features for Radar-Camera 3D Object Detection}
{Kurniawan, I. T. and Trilaksono, B. R.: ClusterFusion: Leveraging Radar Spatial Features for Radar-Camera 3D Object Detection}

\corresp{Corresponding author: Bambang Riyanto Trilaksono (briyanto@lskk.ee.itb.ac.id).}

\begin{abstract}
Thanks to the complementary nature of millimeter wave radar and camera, deep learning-based radar-camera 3D object detection methods may reliably produce accurate detections even in low-visibility conditions. This makes them preferable to use in autonomous vehicles' perception systems, especially as the combined cost of both sensors is cheaper than the cost of a lidar. Recent radar-camera methods commonly perform feature-level fusion which often involves projecting the radar points onto the same plane as the image features and fusing the extracted features from both modalities. While performing fusion on the image plane is generally simpler and faster, projecting radar points onto the image plane flattens the depth dimension of the point cloud which might lead to information loss and makes extracting the spatial features of the point cloud harder. We proposed ClusterFusion, an architecture that leverages the local spatial features of the radar point cloud by clustering the point cloud and performing feature extraction directly on the point cloud clusters before projecting the features onto the image plane. ClusterFusion achieved the state-of-the-art performance among all radar-monocular camera methods on the test slice of the nuScenes dataset with 48.7\% nuScenes detection score (NDS). We also investigated the performance of different radar feature extraction strategies on point cloud clusters: a handcrafted strategy, a learning-based strategy, and a combination of both, and found that the handcrafted strategy yielded the best performance. The main goal of this work is to explore the use of radar's local spatial and point-wise features by extracting them directly from radar point cloud clusters for a radar-monocular camera 3D object detection method that performs cross-modal feature fusion on the image plane.
\end{abstract}

\begin{keywords}
Radar, monocular camera, fusion, 3D object detection, feature extraction, deep learning.
\end{keywords}

\titlepgskip=-15pt

\maketitle

\section{Introduction}
\label{sec:introduction}

\freefootnote{This article has been accepted for publication in IEEE Access. This is the author's version which has not been fully edited and
content may change prior to final publication. Citation information: \href{https://doi.org/10.1109/ACCESS.2023.3328953}{DOI 10.1109/ACCESS.2023.3328953}.

This work is licensed under a Creative Commons Attribution-NonCommercial-NoDerivatives 4.0 License. For more information, see \href{https://creativecommons.org/licenses/by-nc-nd/4.0/}{https://creativecommons.org/licenses/by-nc-nd/4.0/}}

\PARstart{A}{utonomous} vehicles commonly perform 3D object detection to obtain information on the state of their surrounding objects which includes their class, position, orientation, dimensions, and velocity all in 3D. As an accurate and robust object detection performance is required to enable safe navigation, autonomous vehicles often employ multiple sensors of different modalities for the task. The most accurate 3D object detection methods of today are deep learning-based methods that use lidars. Some methods additionally use cameras to support the lidars. While accurate, both lidars and cameras are light-based sensors. As such, the performance of the methods utilizing them alone will degrade in low-visibility conditions such as during rain or nighttime.

Millimeter wave radars generate sparse 2D point clouds on the bird's eye view (BEV) plane with radial velocity and radar cross-section (RCS) measurements that are unique and unobtainable using other sensors. This allows radars to provide information on the objects' motion, shape, size, and material, all of which are valuable for 3D object detection. Moreover, radars utilize radio waves instead of light, making them robust to use in low-visibility conditions. Radars also have a farther detection range than other sensors, enabling them to detect distant objects better. However, the extreme sparsity of radar point clouds makes it hard to perform 3D object detection on radar point clouds alone.

Cameras, on the other hand, provide rich and dense semantic information, enabling accurate object recognition and comprehensive scene understanding. Cameras are especially good at capturing detailed texture, color, and contextual information, all of which are invaluable for perception tasks. However, cameras lack the ability to directly measure depth and velocity, both of which are essential to predict the objects' 3D position, dimensions, and motion. Moreover, cameras do not perform well in low-visibility conditions.

Radar-camera fusion-based methods use radars and cameras together to detect objects, leveraging the complementary strengths of both sensors. The camera provides rich and dense semantic information, complementing the radar's limited semantic information. Meanwhile, the radar provides robust velocity and depth measurements with a farther reach, complementing the camera's limitations in depth and velocity sensing. As the complementary nature of the two sensors addresses their individual limitations, these methods may reliably produce robust and accurate 3D detections even under challenging conditions. Furthermore, the combined cost of the two sensors is cheaper than the cost of a single lidar. Due to these advantages, the millimeter wave radar-camera sensor combination is popular in advanced driving assistance systems (ADAS) \cite{yang2018intelligent}.

However, despite the aforementioned advantages, the difference in characteristics between radars and cameras makes fusing both sensors' information challenging. Additionally, radar-camera fusion-based 3D object detection methods are not as well studied as the lidar-based methods, thus their performance is falling behind \cite{feng2020deep}. Although radars produce point clouds just like lidars, radar point clouds are much sparser and have much lower accuracy and resolution compared to lidar point clouds. Thus, it is hard to directly adapt lidar-based techniques designed for dense and high-accuracy point clouds for radar point clouds.

Generally, fusion-based approaches are categorized into three types according to when the fusion is performed: data-level, feature-level, and decision-level fusion \cite{wei2022mmwave}. Data-level approaches fuse the raw data from both sensor modalities, achieving minimal information loss and enabling them to learn joint features. Even so, they are inflexible, sensitive to misalignment between the sensors, and require a relatively high computational cost. Decision-level approaches fuse the detection results independently acquired from each modality, resulting in flexible and robust methods that have relatively cheap computational costs. However, they suffer from information loss and lack the ability to learn joint features. In the case of radar-camera fusion, it is hard to perform data-level fusion due to the difference in both sensors' characteristics. Decision-level fusion is also unfeasible due to the poor performance of radar-based approaches \cite{svenningsson2021radar, ulrich2022improved}.

\Figure[!tp](topskip=0pt, botskip=0pt, midskip=0pt)[width=0.99\linewidth]{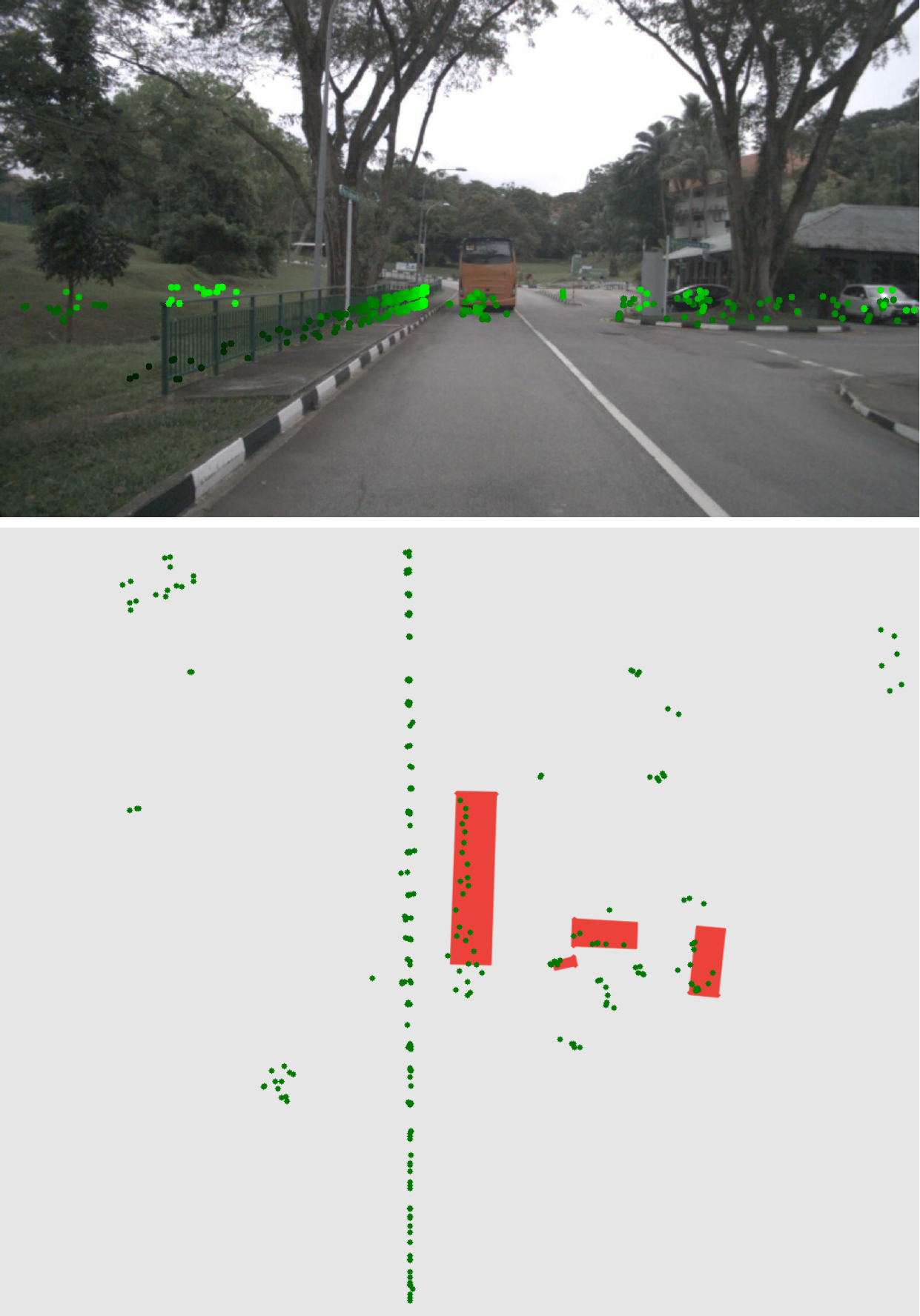}
{A sample radar point cloud from the nuScenes dataset, projected onto the image plane (left) and viewed from above (right). Radar points are shown in green and ground truth object bounding boxes are shown in red. It is simple to make out the dimensions of the objects in the BEV, but relatively harder to do the same in the image perspective view.\label{fig:projecting_bad}}

Feature-level fusion has gained popularity in recent years. It compromises the data-level and the decision-level fusion approaches by fusing the independently extracted features from both modalities. Existing feature-level fusion methods often project the radar point cloud onto another plane, like the BEV plane \cite{kim2020grif} or the camera image plane \cite{chadwick2019distant, nobis2019deep, john2019rvnet, li2020feature, chang2020spatial, nabati2021centerfusion, li2023centertransfuser, long2023radiant}, before treating the projection as an image, extracting features from the projection using techniques developed for image-based object detection, and fusing the extracted features with the image features on the said plane. Projecting the radar point cloud onto the BEV plane preserves the spatial information contained in the point cloud which makes feature extraction easier. However, it requires non-trivial steps to transform the image features to the BEV plane and fuse them with the radar features there. On the other hand, projecting the point cloud onto the image plane is easier and allows for a more straightforward fusion. However, it flattens the depth dimension of the point cloud which is then commonly included in the points' features instead, making it harder to extract the spatial features of the point cloud even though it carries semantic information as well as information on the position, orientation, and dimensions of the objects. This problem is illustrated in Figure \ref{fig:projecting_bad} where we can easily make out the shape, size, and orientation of the objects on the BEV plane while the collapsed depth dimension makes it hard to do the same on the image plane.

Ideally, we would like to be able to extract the radar point cloud's local spatial features effectively while performing feature-level fusion on the image plane. To that end, we present ClusterFusion, a radar-monocular camera 3D object detection architecture that benefits from the simplicity of image plane feature-level fusion while being able to extract and benefit from the radar point cloud's local spatial and point-wise features. ClusterFusion achieves this by first using image-based preliminary 3D object detections to filter and cluster the points in the radar point cloud through a frustum-based association mechanism inspired by CenterFusion \cite{nabati2021centerfusion}. It then extracts features directly from those point cloud clusters without performing any projections, enabling a more effective extraction of the local spatial features. Only after the radar cluster features are extracted they are projected onto the image plane and fused with the image features. The fused features are then fed into the regression heads to generate the final 3D object detections.

To ensure a fair and rigorous comparison, we benchmarked ClusterFusion's performance twice: first against other radar-monocular camera fusion-based methods, and second against methods that use other sensor setups including multiple camera-based methods. Sensor setups that use multiple cameras can resolve depth and use cross-view information, thus boosting the 3D detection performance. Being able to resolve depth is an especially significant help as the performance of monocular methods is mainly limited by poor depth estimation \cite{ma2021delving}. Among all radar-monocular camera methods on nuScenes' \cite{caesar2020nuscenes} test slice, ClusterFusion achieved the highest nuScenes detection score (NDS) of 48.7\% along with the lowest mean orientation (mAOE), velocity (mAVE), and attribute error (mAAE) of 0.424, 0.461, and 0.108 respectively and the second lowest mean translation (mATE) and scale error (mASE) of 0.587 and 0.257 respectively. Compared to methods that use other sensor setups, ClusterFusion managed to achieve a competitive mAAE.

As the proposed architecture operates directly on the radar point cloud, more feature extraction strategies can be explored. In this work, we investigate the effectiveness of different radar feature extraction strategies in extracting radar point cloud cluster features for radar-camera 3D object detection. We implemented and tested the performance of a proposed handcrafted feature extraction strategy, a learning-based strategy based on the KP-CNN architecture which leverages the KPConv operation \cite{thomas2019kpconv}, and a hybrid strategy combining both approaches. The results show that the proposed handcrafted feature extraction strategy achieved the best performance. The preliminary results of our work, discussing the use of radar clusters and handcrafted feature extraction strategy, were presented in \cite{kurniawan2022improving}.

The main contributions of our work are summarized as follows:
\begin{enumerate}
    \item We proposed ClusterFusion, a radar-monocular camera fusion-based 3D object detection method that performs feature extraction directly on radar point cloud clusters, enabling it to benefit from the spatial and point-wise features of the clusters. It performs cross-modal feature fusion on the image plane, making it simpler and faster.
    \item The performance of ClusterFusion is verified on the nuScenes' test slice, where it achieved the highest NDS of 48.7\% among all radar-monocular camera methods on nuScenes' object detection leaderboard.
    \item An investigation on the effectiveness of different radar feature extraction strategies in extracting features from radar point cloud clusters. We found that the proposed handcrafted feature extraction strategy achieved the best performance.
\end{enumerate}

The rest of this paper is organized into six sections. Section \ref{sec:related_works} discusses relevant works and how our work differs from them. Section \ref{sec:method} provides a general overview of our proposed architecture and investigation, which is further detailed in Section \ref{sec:implementation_details}. We discuss the setting and result of our experiments in Section \ref{sec:experiments}. Finally, Section \ref{sec:conclusions} closes this paper with the conclusions.

\section{Related Works}
\label{sec:related_works}

\subsection{Single-Modality 3D Object Detection}
\label{sec:related_works_single_modality}

Lidars are naturally suitable for 3D object detection as it has a 3D sensing capability, full 360-degree range, and high radial and angular accuracy and resolution. As such, lidar-based 3D object detection methods \cite{zhou2018voxelnet, yan2018second, lang2019pointpillars, shi2019pointrcnn, deng2021voxel, yin2021center} were able to produce highly accurate 3D detection with accurate localization and velocity prediction. However, lidars have a shorter detection range and are relatively expensive compared to cameras and radars.

Early camera-based methods \cite{zhou2019objects, simonelli2019disentangling, zhou2020tracking, liu2020smoke, wang2021fcos3d} use a monocular sensor setup, processing only a single frame of image to generate 3D detections. MonoDIS \cite{simonelli2019disentangling} predicts disentangled 3D detection parameters. CenterNet \cite{zhou2019objects, zhou2020tracking}, FCOS3D \cite{wang2021fcos3d}, and some methods inspired by CenterNet \cite{zhang2021objects, ma2021delving, liu2022learning} represent objects as their representative point and regresses 3D detection parameters at those points directly. Consequently, numerous studies have proposed various approaches to improve the monocular depth estimation accuracy \cite{park2021pseudo, wang2022probabilistic, li2022diversity}. DD3D \cite{park2021pseudo} performs depth pre-training supervised by lidar-based depth ground truth information. PGD \cite{wang2022probabilistic} proposed a probabilistic depth representation and leverages geometric prior. MonoDDE \cite{li2022diversity} uses the camera intrinsics and the predicted objects' height to enhance its depth estimation. 

Due to the limitations of the monocular setup, more recent methods \cite{wang2022detr3d, liu2022petr} began to use a multiple-camera setup that can resolve the depth ambiguity. DETR3D \cite{wang2022detr3d} uses a learned set of 3D object queries to index into the multi-view features and generate reference points used to sample 2D features. PETR \cite{liu2022petr} uses object queries that interact with 3D position-aware features generated by encoding 3D position embeddings into the multi-view features.

Recent multiple camera methods \cite{philion2020lift, reading2021categorical, huang2021bevdet, li2022bevformer, li2023bevdepth} extract surround-view features from the multiple images and then transform them into the BEV plane. Methods that use BEV features perform detection on the BEV plane, allowing for better localization accuracy. Lift-Splat-Shoot \cite{philion2020lift}, a vision-based motion planning model, lifts the images into frustum-shaped point clouds and splats them onto the BEV plane according to the images and their camera matrices. BEVDet \cite{huang2021bevdet} leverages the Lift-Splat view transformation strategy with CenterPoint's \cite{yin2021center} heads to perform 3D object detection instead of motion planning. BEVDepth \cite{li2023bevdepth} has found that the performance of the Lift-Splat strategy relies heavily on the depth estimation accuracy. To address this, BEVDepth pairs the Lift-Splat strategy with a more accurate depth estimation module achieved through lidar-based depth supervision and use of camera matrices. To avoid this issue, BEVFormer \cite{li2022bevformer} proposes a different view transformation strategy that leverages predefined grid-shaped BEV queries along with a spatial cross-attention layer to transform features across camera views onto the BEV plane without relying on depth information. Both BEVFormer and BEVDepth also leverage temporal information from multiple frames across timesteps. BEVFormer recurrently fuses past BEV features through a temporal self-attention layer while BEVDepth performs voxel pooling on spatially aligned frustum features from different frames and concatenates the resulting BEV features.

The use of temporal information is becoming increasingly popular for the most recent methods \cite{huang2022bevdet4d, huang2022bevpoolv2, park2023time, yang2023bevformer, han2023exploring, jiang2023polarformer} as it allows camera-based methods to produce highly accurate velocity estimation, among other benefits. Even a simple concatenation of spatially-aligned features from the previous frame, as proposed by BEVDet4D \cite{huang2022bevdet4d} and used by PolarFormer \cite{jiang2023polarformer} and BEVPoolV2 \cite{huang2022bevpoolv2}, can substantially improve the performance of these methods in terms of velocity estimation. The most recent methods such as BEVFormerV2 \cite{yang2023bevformer}, SOLOFusion \cite{park2023time}, and VideoBEV \cite{han2023exploring} explore more complex techniques to better take advantage of long-term temporal information.

Radars are commonly installed in vehicles. However, radars by themselves are not popular for 3D object detection. Due to the low semantic information, the extreme sparsity, and the lack of height information of radar point clouds, performing 3D object detection on radar point clouds alone is a challenging task. Radar-based methods' \cite{svenningsson2021radar, ulrich2022improved} performance on the nuScenes' object detection leaderboard are far inferior to methods that use other sensors, achieving only a seventh of the state-of-the-art method's nuScenes detection score (NDS). 

In general, there are two approaches to radar object detection: grid-based and point-based approaches. Methods of the grid-based approach such as \cite{dreher2020radar, kohler2023improved} rasterize the radar point cloud into a grid-shaped image on the BEV plane that is then processed using image-based object detectors. While simple, the discretization done by this approach might lead to information loss. On the other hand, methods of point-based approach operate and extract features directly from radar point clouds. Radar-PointGNN \cite{svenningsson2021radar} constructs a graph based on the point cloud and employs graph convolutions to extract features. Other works \cite{danzer20192d, tilly2020detection} use PointNets \cite{qi2017pointnet, qi2017pointnet++} to extract features from the point cloud. KPConvPillars and GraphPillars \cite{ulrich2022improved} combine the two approaches by first extracting point-wise features using KPConv \cite{thomas2019kpconv} and graph convolutions respectively before performing grid rendering and processing the grid using a convolution-based pipeline. KPBEVPillars \cite{kohler2023improved}, a grid-based method, also leverages KPConv to better encode local point cloud features during grid rendering.

In this work, we choose to use a monocular 3D object detector based on CenterNet \cite{zhou2019objects} that does not use BEV features or temporal information as ClusterFusion's image-based detector. As we use radar point clouds as input along with camera images, we have access to both depth and velocity measurements, complementing the two main weaknesses of monocular 3D object detection. Compared to methods that use multiple cameras, monocular methods have the advantage of being more affordable, less sensitive to temporal and spatial calibration errors between sensors, and generally require less computation. Moreover, while methods that use BEV features are more accurate and able to localize detections better, they require more computation and hence are generally slower.

In this work, we investigate the effectiveness of KPConv in extracting features from radar point cloud clusters and compare it to other radar feature extraction strategies. KPConv is selected for its superior performance over other learning-based techniques as demonstrated in \cite{ulrich2022improved} and \cite{kohler2023improved}. KPConv is also relatively simple and does not require any graph construction or point grouping.

\begin{figure*}[!htbp]
    \centering
    \includegraphics[width=\textwidth]{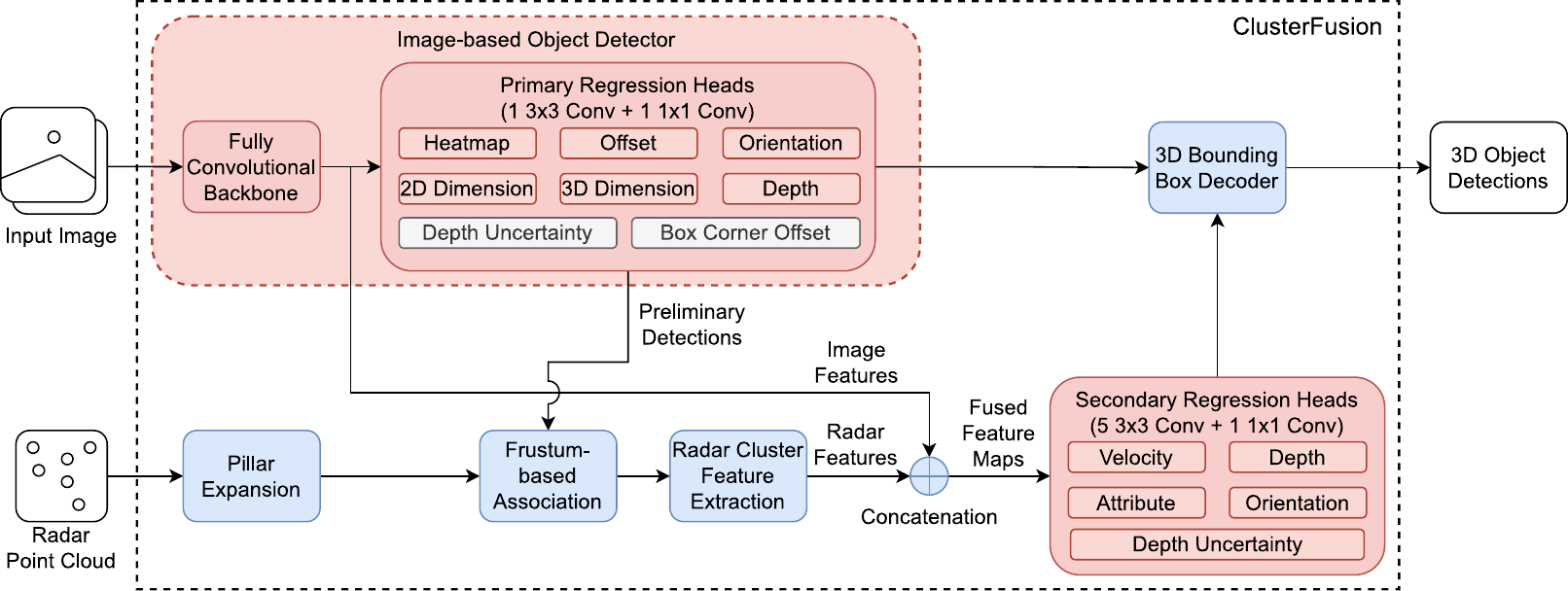}
    \caption{Our proposed architecture, ClusterFusion. Neural network parts are shown in red and non-neural network parts are shown in blue. The regression heads shown in gray act as auxiliary tasks during training and are discarded during inference time.}
    \label{fig:clusterfusion_arch}
\end{figure*}

\subsection{Fusion-based 3D Object Detection}

Fusing camera images with information from point cloud-producing sensors is a popular way to complement cameras' inability to measure depth and achieve better 3D detection performance. Lidars are often paired with cameras due to their ability to measure objects' 3D position with high accuracy as well as the maturity of lidar-based 3D object detection methods. Camera-lidar fusion-based methods \cite{vora2020pointpainting, yoo20203d, pang2020clocs, wang2021pointaugmenting, li2022unifying, yang2022deepinteraction, bai2022transfusion, li2022deepfusion} are able to take advantage of the rich and dense semantic information provided by camera images to complement the highly accurate lidar-based detection, especially at a farther range where the point cloud is sparse.

Despite its relatively low accuracy depth measurements and 2D BEV plane point cloud, radar point clouds offer additional advantages over lidar when fused with camera images. In addition to depth measurements, radar point clouds also have radial velocity measurements and are more robust in low-visibility conditions. Radar-camera 3D object detection methods that use a monocular camera setup were popular in the earlier works \cite{meyer2019deep, kim2020grif, nabati2021centerfusion} and remain actively studied up until today \cite{zhou2023bridging, li2023centertransfuser, long2023radiant}. This line of research aims to use radar information to compensate for the monocular setup's inability to measure both depth and velocity. Meyer and Kuschk \cite{meyer2019deep} adapt the camera-lidar 3D object detector AVOD \cite{ku2018joint} to use with radar point clouds, omitting the feature pyramid network (FPN) part of the architecture. GRIF-Net \cite{kim2020grif} independently produces 3D object proposals from monocular images and radar point clouds and fuses them using a novel gated ROI fusion mechanism. CenterFusion \cite{nabati2021centerfusion} extends the image-based 3D object detector CenterNet \cite{zhou2019objects} to incorporate radar information. CenterFusion generates preliminary 3D object detections using CenterNet, associates the closest radar point inside their frustum-shaped ROI, and uses the associated point's information to refine each detection. CenterTransFuser \cite{li2023centertransfuser} proposes an attention-based cross-transformer module to facilitate deep radar-camera interaction and information fusion along with a depth-adaptive threshold filtering method to filter out noisy radar detections for more accurate association. RADIANT \cite{long2023radiant} improves radar-camera association by explicitly predicting the position offset between each radar point and the objects' 3D center. RADIANT also proposes a modular radar-camera fusion architecture that can improve existing monocular methods' depth estimation accuracy without retraining. RCBEV \cite{zhou2023bridging} uses a view transformer module to lift the image features onto the BEV plane where it performs a two-stage radar-camera fusion. RCBEV uses a strategy based on VoxelNet \cite{zhou2018voxelnet} and ConvLSTM \cite{shi2015convolutional} to extract radar features which will be used to augment the image BEV features. It also generates object heatmaps from the radar point clouds which will be used to augment the object heatmaps obtained from the fused features.

Recently, radar-multiple camera fusion-based 3D object detection methods \cite{wu2023mvfusion, pang2023transcar, kim2023craft} have gained popularity. MVFusion \cite{wu2023mvfusion} leverages a cross-attention mechanism-based radar-guided fusion transformer to fuse semantically aligned radar features with the features obtained from multiple images. CRAFT \cite{kim2023craft} follows a similar outline to CenterFusion but it uses a soft polar association module that associates radar points with object proposals using uncertainty-aware thresholds in the polar coordinate and then uses a spatio-contextual fusion transformer that allows spatial and contextual information exchange between the image and radar features to adaptively fuse them. The most recent radar-camera 3D object detection methods \cite{klingner2023x3kd, stacker2023rc, kim2023rcm, kim2023crn, lei2023hvdetfusion} additionally use BEV image features on top of using multiple cameras. CRN \cite{kim2023crn} and RCM-Fusion \cite{kim2023rcm} address the inaccurate transformation of image features onto the BEV plane by using radar information in the view transformation process. CRN additionally leverages an attention-based multi-modal feature aggregation module to handle spatial misalignment in fusing the feature maps. HVDetFusion \cite{lei2023hvdetfusion}, the current state-of-the-art radar-multiple camera method, uses a modified BEVDet4D \cite{huang2022bevdet4d}, which benefits from temporal information, to extract BEV image features and produce preliminary detections which are used to filter noisy radar points. The features extracted from the filtered radar point cloud are in turn used to augment the BEV image features which are used to produce the final 3D object detections.

In this work, we tackle the radar-monocular image 3D object detection problem. As stated in Section \ref{sec:related_works_single_modality}, the radar-monocular camera setup is chosen for its affordability, robustness to calibration errors, and simplicity. The radar complements the monocular setup's inability to sense depth and velocity, theoretically eliminating the need to use multiple cameras and temporal information. We perform cross-modal feature fusion on the image plane as it is generally simpler and faster than performing fusion on the BEV plane. Previous works that perform feature fusion on the image plane project the radar point clouds onto the image plane and extract features from the projection as if it was an image \cite{chadwick2019distant, nobis2019deep, john2019rvnet, li2020feature, chang2020spatial, nabati2021centerfusion, li2023centertransfuser, long2023radiant}. This flattens the point clouds' depth dimension and makes it harder to extract the local spatial features of the point clouds. Our work differs from previous works in how we handle the radar point clouds. We propose to form clusters from the radar point clouds and extract features from the clusters directly in their point cloud form before performing any projections, enabling an easier extraction and utilization of the radar point clouds' local spatial and point-wise features.

\section{Method}
\label{sec:method}

ClusterFusion, our proposed architecture illustrated in Figure \ref{fig:clusterfusion_arch}, is inspired by CenterFusion \cite{nabati2021centerfusion} and adopts its overall design. The architecture takes a monocular image along with a radar point cloud as input and performs 3D object detection in two stages. The first stage generates preliminary 3D object detections from the input image. The second stage leverages the radar's BEV position and radial velocity measurements to refine the preliminary detections' velocity, depth, orientation, and attribute prediction. 

In the first stage, the image is passed through an image-based 3D object detector, consisting of a fully convolutional backbone and a set of primary regression heads. We obtain preliminary 3D object detections from the primary regression heads and retain the image feature maps generated by the fully convolutional backbone. The radar point cloud information is not utilized in this stage.

To leverage information from the input radar point cloud, we first address the lack of height information on the radar points and the low angular resolution of the radar sensor by performing pillar expansion, expanding each radar point into a fixed-size pillar in 3D space. The radar points from the preprocessed point cloud are then clustered and associated with the preliminary detections from the first stage through our proposed frustum-based association mechanism. This mechanism filters out clutters, forms radar point cloud clusters, and associates a radar point cloud cluster to each preliminary detection at the same time, facilitating the extraction of both local spatial features and point-wise features of the point cloud clusters. We apply a radar feature extraction strategy on the radar point clusters to obtain the radar feature maps. In this paper, we use three different alternatives of radar feature extraction strategies to extract the clusters' features and compare their performance.

In the second stage, our architecture performs feature-level fusion by concatenating the image feature maps obtained in the first stage with the radar feature maps to create fused feature maps. The secondary regression heads then use these fused feature maps to produce an improved prediction of the preliminary detections' velocity, depth, orientation, and attribute. Finally, the detection parameters predicted by the primary and secondary regression heads are combined and decoded by the 3D bounding box decoder to obtain the final 3D object detection bounding boxes. It's worth noting that our architecture does not generate object detections from the input radar point cloud alone. Instead, it fuses individually extracted feature maps from both the camera and the radar. As our architecture relies on a deep learning-based image-based 3D object detector and regression heads to handle the radar point clouds and to produce 3D object detections, our architecture is data-driven by nature.

\subsection{Image-based 3D Object Detector}

We use an image-based 3D object detector based on CenterNet \cite{zhou2019objects} with a 34-layer modified Deep Layer Aggregation (DLA-34) \cite{yu2018deep} network, also proposed in \cite{zhou2019objects}, as its backbone. The image-based detector is responsible for carrying out the first stage of detection by producing preliminary 3D detections from the input image. As with CenterNet, our image-based detector represents objects as individual representative points. To detect objects, our image-based detector feeds the image features extracted by the backbone to the primary regression heads to predict a heatmap of the representative points along with a set of 3D detection parameters. The values of the heatmap represent the confidence of an object belonging to a certain class being in each position. The 3D detection parameters are the offsets from the representative point to the projected 3D center as well as the objects' 3D dimensions, depths, and yaw-axis orientations. The predicted heatmap of representative points and 3D detection parameters are then decoded to obtain the preliminary 3D object detections.

In addition to the necessary parameters to produce 3D object detections, the detector also learns to predict the objects' 2D bounding box dimensions, the depth estimation uncertainties, and the 2D offsets between the projected centers and corners of the 3D bounding boxes as auxiliary tasks to help to learn shared features during training. As the latter two parameters are not used outside training, their corresponding regression heads are discarded during inference time. On the other hand, the objects' 2D bounding box dimensions are kept as they are needed in the frustum-based association mechanism.

We applied techniques proposed by previous works to improve the performance of our image-based detector. These techniques include adding several regression heads, some of which are briefly discussed above, adjusting some of the existing regression heads, using objects' projected 3D bounding box centers as representative points, and decoupling the detection of truncated objects from other objects.

\subsubsection{Primary Regression Heads}

We added several heads to the primary regression heads and introduced changes to some of the other heads. Inspired by MonoFlex \cite{zhang2021objects}, we added the depth uncertainty head to predict the aleatoric uncertainties \cite{kendall2017uncertainties}, specifically the log standard deviation, of the depth estimate. This regression head predicts the uncertainties that might arise due to the randomness inherent to the data such as sensor noises, variable lighting, surface textures, depth discontinuities, and scene ambiguities. During training, the predicted variance is incorporated into the depth regression head's loss function described in Equation \ref{eq:depth_loss}, helping supervise the learning process and enabling the head to learn more from low-uncertainty samples and less from high-uncertainty samples. The aforementioned loss function is described and discussed further in Section \ref{sec:training_losses}. 

As with CenterFusion's \cite{nabati2021centerfusion} depth head, our depth head predicts the depths transformed using the inverse sigmoidal transform $\hat{d}_{\text{sig}}$ introduced in \cite{eigen2014depth} instead of predicting objects' depth $\hat{d}$ directly. Equation \ref{eq:dep_sigmoid} below describes how $\hat{d}$ relates to $\hat{d}_{\text{sig}}$
\begin{equation}
    \hat{d} = \frac{1}{\sigma(\hat{d}_{\text{sig}})} - 1,
\label{eq:dep_sigmoid}
\end{equation}
with $\sigma(\cdot)$ being the sigmoid function. Learning to predict $\hat{d}_{\text{sig}}$ is easier than $\hat{d}$ due to it having a narrower range of values.

To help improve the overall detector performance, we added the box corner offset head to learn the auxiliary task of predicting the 2D offsets between the projected bounding box 3D center and corners as proposed by MonoCon \cite{liu2022learning}. As this regression head is only used to help learn shared features during training, this regression head is discarded during inference time. We also opted to change the object's 2D dimensions formulation. Instead of predicting the width and height of the bounding boxes, the distance from the objects' representative points to the left, top, right, and bottom sides of the bounding boxes are predicted. This change removes the constraint of having to use the objects' 2D bounding box center as the objects' representative point.

All the primary regression heads are formed by a 3$\times$3 convolutional layer followed by a 1$\times$1 convolutional layer. We use LeakyReLU as the primary regression heads' activation function, save for the representative point heatmap head for which we use the sigmoid function instead. The activation function is applied after each convolutional layer.

\subsubsection{Objects' Projected 3D Center as Representative Point}

MonoDLE \cite{ma2021delving} has found that localization error is the key factor that limits the accuracy of monocular 3D object detection methods. They have also found that the localization error can be reduced by using the objects' projected 3D center as their representative point instead of the objects' 2D bounding box center as in the original CenterNet. Thus, we follow MonoDLE to use the objects' projected 3D center as their representative point. It might be important to note that even if the detector uses the objects' projected 3D center as their representative point, the offset regression head is still needed to predict the subpixel offsets between the predicted projected 3D centers that are quantized and the ground truth projected 3D centers that are calculated directly from the objects' ground truth 3D position and the camera matrices.

\subsubsection{Decoupling the Detection of Truncated Objects}

Truncated objects are challenging to detect and localize well as their projected 3D center might lie outside of the image. To improve the detection and localization accuracy of truncated objects, we follow MonoFlex \cite{zhang2021objects} to decouple the detection of truncated objects from other objects. If a truncated object's projected 3D center lies outside of the image, then the detector will not use the projected 3D center as its representative point. Instead, the detector will use the intersection point between the image's edge and the line connecting the object's 2D bounding box center to the projected 3D center as the object's representative point. To recover the position of the projected 3D center, the offset regression head will predict the offset between the new representative point and the projected 3D center. To improve the detection performance for truncated objects whose representative point is on the image's edge, we adopted the edge fusion module proposed by MonoFlex designed to enhance the detector's ability to learn and extract features on the image's edge. We do not use MonoFlex's depth from keypoints and depth ensemble strategy as we have the radar measurements to help estimate objects' depth.

\subsection{Frustum-based Association}

We propose a frustum-based association mechanism inspired by CenterFusion \cite{nabati2021centerfusion} to filter out clutters, form radar point cloud clusters, and associate radar point cloud clusters to the preliminary 3D detections generated by the image-based detector. The mechanism, illustrated in Figure \ref{fig:frustum_assoc}, works by generating a frustum-shaped ROI for each preliminary detection using its 2D bounding box, estimated depth, 3D dimensions, and orientation. Radar points outside of the ROI frustum of an object are of no interest and are filtered out. The ROI frustum generation part of the association mechanism itself is identical to the one used in CenterFusion.

\Figure[!tp](topskip=0pt, botskip=0pt, midskip=0pt)[width=0.99\linewidth]{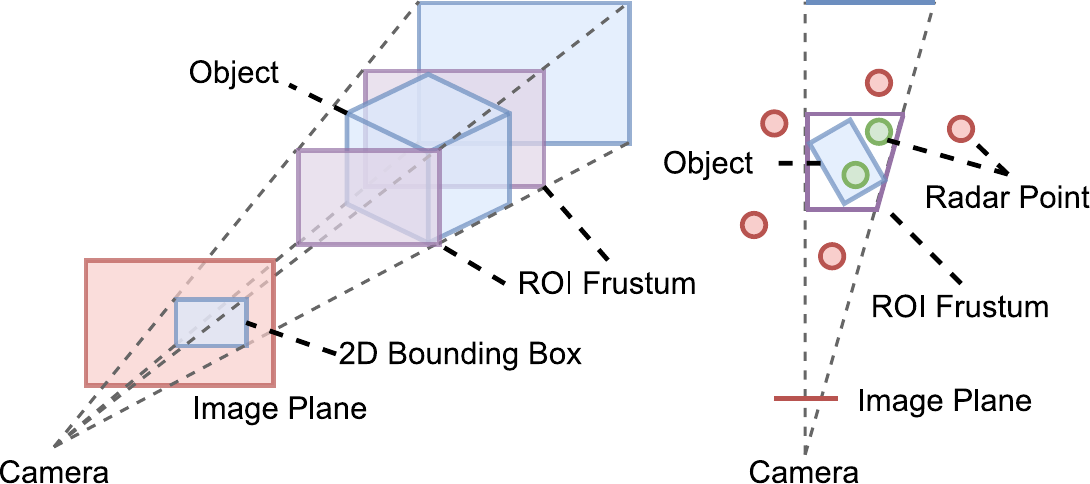}
{An illustration of the frustum-based association step, displayed in perspective (left) and top view (right). The radar points inside the ROI frustum (green) are associated with the corresponding object while other radar points (red) are filtered out.\label{fig:frustum_assoc}}

Once we have the ROI frustum, instead of following CenterFusion to associate the single closest radar point to the ego-vehicle, we associate every radar point inside an ROI frustum with the preliminary detection that produced the said ROI frustum. The proposed frustum-based association mechanism aims to associate each object with a radar cluster containing every point produced by the object. Thus, not only does the frustum-based association mechanism serve to match radar points to preliminary detections, but it also serves to group and cluster the radar points in the point cloud into local neighborhoods. This way, we would be able to extract the spatial features contained in the geometry of the clusters and also learn how the point-wise features of the points in a cluster interact with each other, both of which will be impossible if we associate only a single radar point to each preliminary detection or if we project the radar point cloud directly to the image plane.

\subsection{Radar Cluster Feature Extraction}
\label{sec:method_radar_cluster_feature_extraction}

The radar feature extraction module is responsible for extracting radar features from the radar point clusters formed by the frustum-based association mechanism. The extracted radar features will then be concatenated with the image features extracted by the image-based detector's backbone to get the fused feature maps. In this work, we use three different radar feature extraction strategies to fill the role of the radar feature extraction module. The first strategy is a handcrafted feature extraction strategy that simply takes the statistical quantities of the point-wise feature value distribution among all the points in a cluster. The second strategy is a learning-based strategy that uses a model with an architecture based on KP-CNN that leverages the KPConv operator \cite{thomas2019kpconv} to learn the features of a cluster. Lastly, the third strategy employs both aforementioned strategies in parallel and concatenates the feature maps extracted by both strategies.

\subsubsection{Handcrafted Radar Feature Extraction Strategy}

Handcrafted feature extraction strategies are desirable as they do not require any additional learnable model parameters and training processes. Handcrafted features are also relatively cheap to compute. However, the effectiveness of handcrafted feature extraction strategies highly depends on the quality of the features' formulation. As shown in Figure \ref{fig:projecting_bad}, the geometry, namely the shape, size, and orientation as well as the velocity of the radar clusters can serve as a rough estimate of the state of the object that produces it. Thus, we would want a radar cluster feature formulation that can capture the geometry and the aggregate velocity of a cluster. 

As the statistical quantities of the radar points' 2D position distribution can describe a cluster's geometry, it makes sense to use these quantities as the cluster's features. Likewise, the statistical quantities of the radar points' radial velocity distribution can be used as a cluster's features as they describe the aggregate velocity of the cluster. We use the maximum and minimum values as well as the mean of the normalized 2D position $\mathbf{x} = \left(x, y\right)$ and the ego-motion-compensated 2D-projected radial velocity $\mathbf{v} = \left(v_x, v_y\right)$ of the radar points inside a cluster as its features.

To help capture a cluster's general orientation, we perform line fitting onto the cluster, convert the slope of the best-fitting line to orientation in radians, and use the orientation as an additional feature. Given a radar cluster consisting of $N$ points, the slope of the best-fitting line $m$ is calculated in closed-form using the linear least-squares method given by Equation \ref{eq:ort_lls} below
\begin{equation}m = \frac{\sum^N_{n=1}(x_n - \bar{X})(y_n - \bar{Y})}{\sum^N_{n=1}(x_n - \bar{X})^2},\label{eq:ort_lls}\end{equation}
with $x_n$ and $y_n$ being the normalized $x$ and $y$-axis position of the $n$-th point, and $\bar{X}$ and $\bar{Y}$ the mean normalized $x$ and $y$-axis position of the points. With this feature, we have a set of 13 features $\mathbf{f}$ for each radar cluster described in Equation \ref{eq:handcrafted_feats} below
\begin{equation}
    \begin{split}
        \mathbf{f} &= \left( \mathbf{f}_{\text{max}}, \mathbf{f}_{\text{min}}, \mu_\mathbf{f}, m \right), \\
        \mathbf{f}_{\text{max}} &= \left( x_{\text{max}}, y_{\text{max}}, v_{x,\text{max}}, v_{y,\text{max}} \right), \\
        \mathbf{f}_{\text{min}} &= \left( x_{\text{min}}, y_{\text{min}}, v_{x,\text{min}}, v_{y,\text{min}} \right), \\
        \mu_\mathbf{f} &= \left( \mu_{x}, \mu_{y}, \mu_{v_{x}}, \mu_{v_{y}} \right),
    \end{split}
\label{eq:handcrafted_feats}
\end{equation}
with $(\cdot)_{\text{max}}$, $(\cdot)_{\text{min}}$, and $\mu_{(\cdot)}$ being the maximum, minimum, and mean value of the quantity $(\cdot)$ among all radar points inside the cluster.

\Figure[!tp](topskip=0pt, botskip=0pt, midskip=0pt)[width=0.99\linewidth]{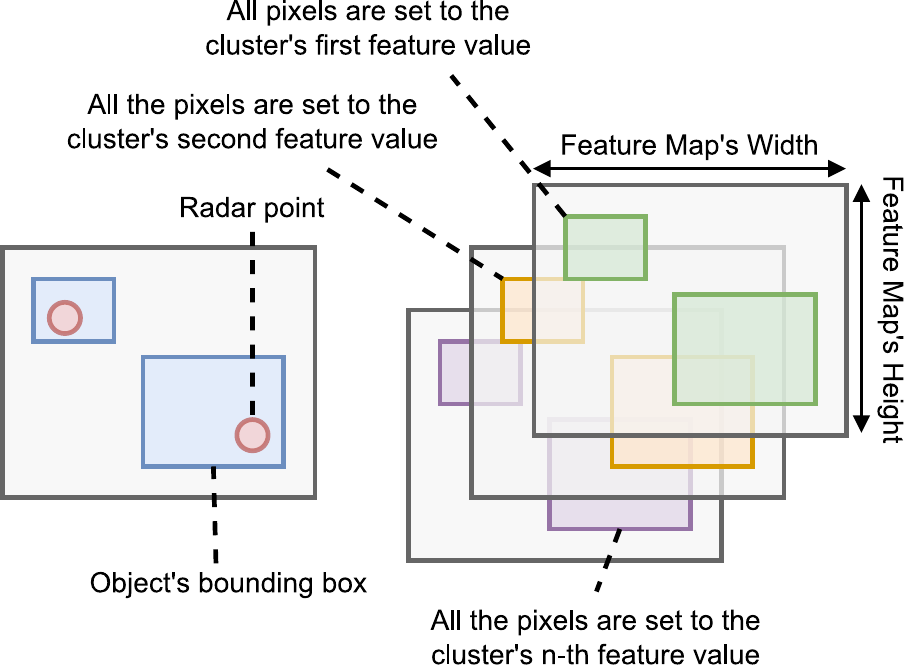}
{An illustration of the radar cluster features-to-heatmap conversion.\label{fig:heatmap_conv}}

To transform the extracted radar cluster features onto the image plane and enable fusion with the image feature maps, we use a feature-to-heatmap conversion mechanism illustrated in Figure \ref{fig:heatmap_conv}, similar to the one used by CenterFusion. As we have 13 cluster features, the resulting image-plane heatmap will have 13 channels with each channel containing the value of a different cluster feature. For each cluster, we set the values of every pixel inside the 2D bounding box of its corresponding preliminary detection to the value of the cluster features, over all the heatmap channels. Thus, we obtain a 13-channel heatmap that contains the features of all clusters in the scene and is ready to be concatenated with the image features.

\subsubsection{Learning-based Radar Feature Extraction Strategy}

In contrast to the handcrafted strategy, the learning-based feature extraction strategy requires additional learnable model parameters and training processes. Moreover, it is also relatively expensive to compute. However, it is able to learn the optimal features from the radar clusters on its own. We use a KPConv-based \cite{thomas2019kpconv} model with an architecture based on KP-CNN \cite{thomas2019kpconv} as our radar feature extractor. We follow KP-CNN's architecture up until its global average pooling layer, skipping the final fully connected and softmax layer that is used to generate class predictions. The overall radar feature extractor architecture is shown in Figure \ref{fig:kp-cnn_arch}.

\Figure[!tp](topskip=0pt, botskip=0pt, midskip=0pt)[width=0.99\linewidth]{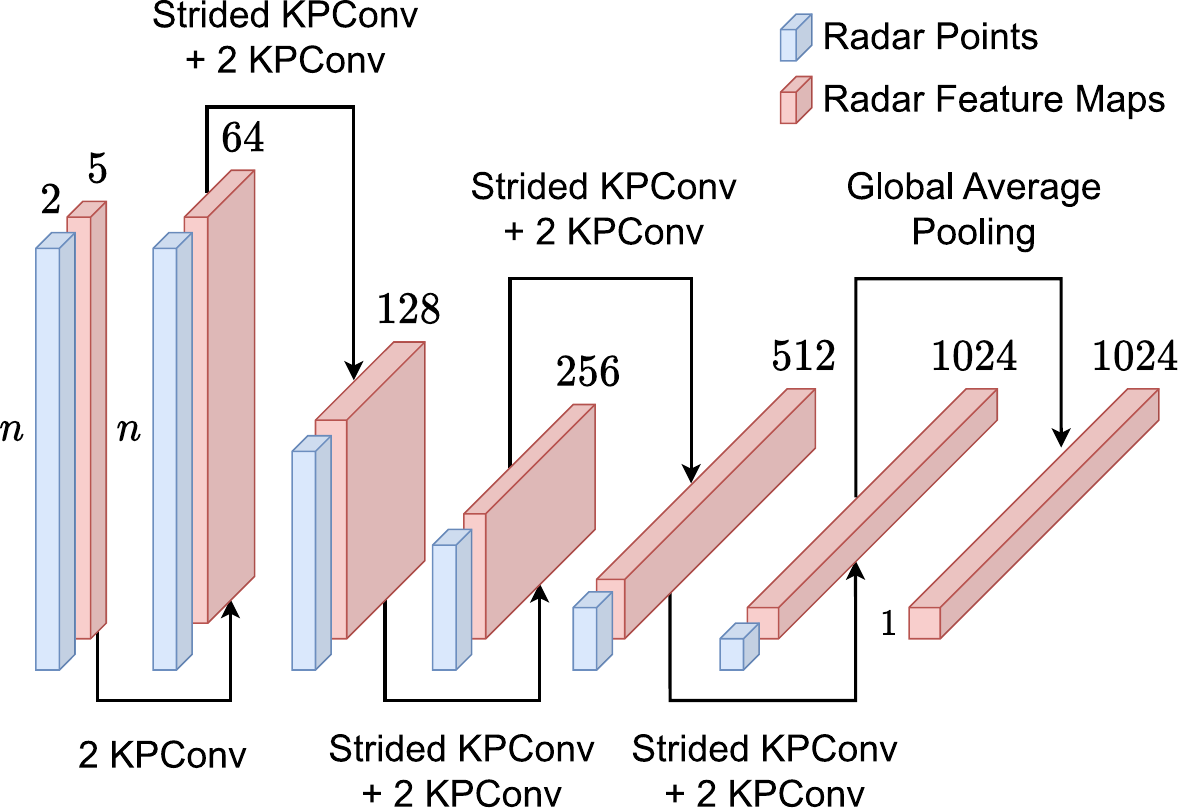}
{The KPConv-based radar feature extractor architecture.\label{fig:kp-cnn_arch}}

We used each radar point's normalized 2D position $\mathbf{x}$, ego-motion-compensated 2D-projected radial velocity $\mathbf{v}$, and a constant feature with the value of 1.0 as each point's individual features, amounting to a total of 5 features per point. The constant feature is added to encode the local geometric characteristics of the points in the cluster through interaction with the KPConv kernel points' positions and weights. Its constant value across all points and time ensures that every point's local geometric characteristics are consistently captured and encoded regardless of the specific value or variations of the point's other features. The radar feature extractor consists of 5 KPConv-based convolutional layers. Each layer after the first layer performs strided convolution, thus reducing the number of points of interest while increasing the dimension of the radar feature maps. The radar feature extractor will process each radar cluster produced by the frustum-based association mechanism and produce a single point with a 1024-channel feature map that represents the radar cluster. The feature maps of this point will then be taken as the corresponding cluster's features. It might be important to note that the strided KPConv-based convolution requires both the grid subsampling and fixed radius near neighbor search operations that are relatively computationally expensive.

To transform the radar cluster features onto the image plane and prepare it for fusion with the image feature maps, we use a mechanism identical to the one used in the handcrafted radar feature extraction strategy, shown in Figure \ref{fig:heatmap_conv}. However, as we now have a 1024-channel feature map, the resulting heatmap will have 1024 channels with each channel corresponding to a different feature map channel.

\subsection{Secondary Regression Heads}

Given the fused feature maps obtained by concatenating the image features extracted by the image-based detector's backbone and the radar features extracted by the radar feature extractor, the secondary regression heads will produce refined predictions of the objects' velocity, depth, yaw-axis orientation, attribute, and depth estimation uncertainty. Unlike the depth uncertainty head from the primary regression heads, the secondary depth uncertainty head is not discarded during inference time as the refined depth estimation uncertainty predictions are used to calculate the final 3D detection confidence in the decoder. The rest of the refined predictions are passed to the decoder along with the predictions made by the primary regression heads to be decoded into the final 3D object detections. 

The secondary regression heads are responsible for learning the joint representations and cross-modality interactions between the radar and image features after the concatenation. Thus, the secondary regression heads are designed to be more complex than the primary heads, each consisting of five 3$\times$3 convolutional layers followed by a 1$\times$1 convolutional layer. As with the primary heads, the secondary heads also use LeakyReLU as their activation function, applied once right before and right after the 1$\times$1 convolutional layer.

\subsection{3D Bounding Box Decoder}

To decode the predictions made by the regression heads into 3D object detections, the decoder takes the $K$ points with the highest confidence across all classes from the heatmap of representative points produced by the primary regression heads. This is equivalent to taking the $K$ points with the highest heatmap value across all classes. The decoder then assigns each candidate its 3D center position, 3D dimensions, orientation, velocity, and attribute from the corresponding predictions according to the candidate's position on the heatmap. The 3D center position itself is calculated from the predicted projected 3D center, depth, as well as the camera's intrinsic and extrinsic matrices.

The decoder incorporates the refined depth estimation log standard deviation $\log\left(\sigma\right)$ predicted by the secondary depth uncertainty head into the final 3D detection confidence $p_{\text{3D}}$ following MonoDDE \cite{li2022diversity} and MonoPixel \cite{kim2022boosting}. By doing so, the confidence of candidates with high depth estimation uncertainty can be attenuated, thus filtering out candidates that are poorly localized. We use the formulation proposed in MonoPixel to calculate the depth estimation confidence $p_{\text{dep}}$ given by Equation \ref{eq:depth_confidence} as follows
\begin{equation}p_{\text{dep}} = e^{-\sigma^2}.\label{eq:depth_confidence}\end{equation}
The depth estimation confidence can then be used to weigh the class confidence $p_{k}$, taken from the candidates' representative point heatmap value, following Equation \ref{eq:3d_confidence} below
\begin{equation}p_{\text{3D}} = p_{\text{dep}} p_{k}.\label{eq:3d_confidence}\end{equation}
Candidates with sufficient 3D detection confidence will be taken as final 3D object detections while the rest are filtered out.

\subsection{Training Losses}
\label{sec:training_losses}

We utilize different training losses to train each regression head according to the task the head does. In the equations, we denote predicted quantities with a hat while their ground truth counterparts are written without a hat. The focal loss \cite{lin2017focal} is used to train both the representative point heatmap head and the attribute head as it upweights difficult samples such as occluded or rare objects and downweights easy samples, helping to improve detection and classification accuracy on difficult and rare classes. The classification loss $L_{\text{cls}}$ for the two heads is given by the following Equation \ref{eq:focal_loss} below
\begin{equation}
\begin{split}
    &L_{\text{cls}} = -\frac{1}{M} \sum_{xyc} \\
    &\begin{cases} 
        (1 - \hat{Y}_{xyc})^\alpha \log(\hat{Y}_{xyc}), & Y_{xyc} = 1 \\
        (1 - Y_{xyc})^\beta (\hat{Y}_{xyc})^\alpha \log(1 - \hat{Y}_{xyc}), & \text{otherwise},
    \end{cases}
\end{split}
\label{eq:focal_loss}
\end{equation}
where $M$ is the number of objects, $(x, y, c)$ the coordinates of the pixel on the heatmap, $Y_{xyc}$ the confidence value at the given heatmap coordinate, $\alpha$ and $\beta$ the hyperparameter of the focal loss. The values of the hyperparameters are chosen to be $\alpha = 2$ and $\beta = 4$ following CenterNet \cite{zhou2019objects}.

Following MonoFlex \cite{zhang2021objects}, the offset regression head that predicts the offset between the objects' representative point and projected 3D center is trained using the log-scale L1 loss for truncated objects and the normal L1 loss for the other objects. The log-scale L1 loss is used for truncated objects as it is more robust against outliers than the normal L1 loss. The formulation of the offset loss $L_{\text{off}}$ is described in Equation \ref{eq:offset_loss} below
\begin{equation}
    L_{\text{off}} = \begin{cases}
    \frac{1}{M} \sum^{M}_{k=1} \log(1 + | \mathbf{o}_k - \hat{\mathbf{o}}_k |), & \text{truncated} \\
    \frac{1}{M} \sum^{M}_{k=1} | \mathbf{o}_k - \hat{\mathbf{o}}_k |, & \text{otherwise},
    \end{cases}
    \label{eq:offset_loss}
\end{equation}
where $\mathbf{o}_k = (o_{x,k}, o_{y,k})$ is the aforementioned offset.

The velocity, 3D dimensions, and box corner offset heads predicting the velocity $\mathbf{V}$, 3D dimensions $\mathbf{D}_{\text{3D}}$, as well as the 2D offsets between the box projected corners and projected 3D center $\mathbf{c}$ are performing simple regression. The L1 loss is used as it is more robust to outliers than its alternatives such as the L2 loss. The loss function for each head is described in Equation \ref{eq:vel_loss}, \ref{eq:dim3d_loss}, and \ref{eq:corner_loss} below
\begin{equation}
    L_{\text{vel}} = \frac{1}{M} \sum^{M}_{k=1} | \mathbf{V}_k - \hat{\mathbf{V}}_k |, \ \mathbf{V} = (V_x, V_y),
\label{eq:vel_loss}
\end{equation}
\begin{equation}
\begin{split}
    L_{\text{dim3D}} &= \frac{1}{M} \sum^{M}_{k=1} | \mathbf{D}_{\text{3D}, k} - \hat{\mathbf{D}}_{\text{3D}, k} |, \\ 
    \mathbf{D}_{\text{3D}} &= (D_{\text{width}}, D_{\text{length}}, D_{\text{height}}),
\end{split}
\label{eq:dim3d_loss}
\end{equation}
\begin{equation}
    L_{\text{corner}} = \frac{1}{M} \sum^{M}_{k=1} | \mathbf{c}_k - \hat{\mathbf{c}}_k |, \ \mathbf{c} = (c_x, c_y), \ \mathbf{c} \in \text{corners}.
\label{eq:corner_loss}
\end{equation}

The orientation heads predicting the objects' yaw-axis orientation are trained using the MultiBin loss \cite{mousavian20173d} that divides the orientation range into $N_{\theta}$ bins, which in our case $N_{\theta} = 4$. For each bin, the orientation heads predict the confidence that the orientation falls into the bin $\hat{b}$ as well as the sine and cosine of the orientation offset from the center of the bin $\hat{\sin}(\Delta\theta_{i})$ and $\hat{\cos}(\Delta\theta_{i})$. Described in Equation \ref{eq:rotcls_loss} below is the orientation bin classification loss $L_{\text{rotcls}}$ used to train the confidence prediction part of the orientation heads
\begin{equation}
    L_{\text{rotcls}} = -\frac{1}{M} \sum^{M}_{k=1} \frac{1}{N_{\theta}} \sum^{N_{\theta}}_{i=1} L_{\text{BCE}} \left( \hat{b}_{k, i}, b_{k, i} \right),
\label{eq:rotcls_loss}
\end{equation}
where $L_{\text{BCE}}$ is the binary cross-entropy (BCE) loss. On the other hand, the orientation offset prediction part of the orientation heads is trained using the bin residual loss $L_{\text{rotres}}$ described in Equation \ref{eq:rotres_loss} as follows
\begin{equation}
\begin{split}
    L_{\text{rotres}} &= -\frac{1}{M} \sum^{M}_{k=1} \frac{1}{n_{\theta}} \sum^{n_{\theta}}_{i=1} \\
     &L_{\text{L1}} \left( \hat{\cos}(\Delta\theta_{k, i}), \cos(\theta_{k, i} - c_i) \right) \\
    + &L_{\text{L1}} \left( \hat{\sin}(\Delta\theta_{k, i}), \sin(\theta_{k, i} - c_i) \right),
\end{split}
\label{eq:rotres_loss}
\end{equation}
where $n_{\theta}$ is the number of bins covering the orientation, $L_{\text{L1}}$ the L1 loss function, $\theta_{k, i}$ the ground truth orientation offset for the $k$-th object and $i$-th bin, and $c_i$ the center of the $i$-th bin. $L_{\text{rotcls}}$ and $L_{\text{rotres}}$ are summed to get the rotation loss $L_{\text{rot}}$ described in Equation \ref{eq:rot_loss} below
\begin{equation}
    L_{\text{rot}} = L_{\text{rotcls}} + L_{\text{rotres}}.
\label{eq:rot_loss}
\end{equation}

The depth and depth uncertainty heads predicting the objects' depth $d$ and estimation log standard deviation $\log\left(\sigma\right)$ are trained using an uncertainty-attenuated L1 loss $L_{\text{dep}}$ proposed by \cite{kendall2017uncertainties} described in Equation \ref{eq:depth_loss} as follows
\begin{equation}
    L_{\text{dep}} = \frac{1}{M} \sum^{M}_{k=1} \left[ \frac{|d_k - \hat{d}_k|}{\hat{\sigma}^2_k} + \log \hat{\sigma}^2_k \right].
\label{eq:depth_loss}
\end{equation}
This loss function supervises both the depth and depth uncertainty heads at the same time. As predicting the estimation variance might cause numerical issues, we opted to have the depth uncertainty head predict the log standard deviation $\log\left(\sigma\right)$ instead. Additionally, this simplifies the loss function calculation by replacing the need to calculate logarithms with exponential functions.

The generalized intersection-over-union (GIoU) proposed by \cite{rezatofighi2019generalized} is used to supervise the 2D dimensions head predicting the dimensions of the 2D bounding box $\mathbf{D}_{\text{2D}}$. The GIoU loss, designed to supervise 2D bounding box regression, takes into account the shape and size of the bounding boxes which makes it more effective and robust than the L1 loss. The 2D dimensions loss $L_{\text{dim2D}}$ is described in Equation \ref{eq:dim2d_loss} below
\begin{equation}
\begin{split}
    L_{\text{dim2D}} &= \frac{1}{M} \sum^{M}_{k=1} \text{GIoU}(\hat{\mathbf{D}}_{\text{2D}, k}, \mathbf{D}_{\text{2D}, k}), \\
    \mathbf{D}_{\text{2D}} &= (D_{\text{left}}, D_{\text{top}}, D_{\text{right}}, D_{\text{bottom}}).
\end{split}
\label{eq:dim2d_loss}
\end{equation}

All the aforementioned loss functions are summed together to get the total loss function $L_{\text{tot}}$, defined in Equation \ref{eq:total_loss} as follows
\begin{equation}
\begin{split}
    L_{\text{tot}} &= L_{\text{cls}} + L_{\text{off}} + L_{\text{vel}} + L_{\text{dim3D}} + L_{\text{rot}} \\
    &+ L_{\text{dep}} + 0.1 L_{\text{dim2D}} + 0.5 L_{\text{corner}},
\end{split}
\label{eq:total_loss}
\end{equation}
with $L_{\text{dim2D}}$ and $L_{\text{corner}}$ weighted less than the other losses as both of them are auxiliary training losses that are only used to help the model learn shared features during training.

\section{Implementation Details}
\label{sec:implementation_details}

The model is implemented in Python using the PyTorch library based on the source code provided by CenterFusion \cite{nabati2021centerfusion}. We reimplemented various parts of the existing source code such as the frustum-based association and heatmap conversion to take advantage of vectorization and enable faster computations. Our implementation of CenterFusion can be executed over 3 times faster than the original implementation, as shown in Table \ref{tab:runtime_comparison} and discussed in Section \ref{subsec:experiments_sota_comp}. We implemented ClusterFusion on top of our implementation of CenterFusion.

\subsection{Architecture Details}

The model takes single frames of 800$\times$448 pixels-sized monocular images along with radar point clouds accumulated from the latest 6 sweeps as input. The model's pillar expansion module expands each radar point in the point cloud into a 0.2 meters long, 0.2 meters wide, and 1.5 meters tall pillar in the 3D space. Radar points closer than 1.0 meters and farther away than 60.0 meters are discarded. The model decodes $K = 100$ object candidates from each input image frame-radar point cloud pair.

\subsection{Training Details}

We trained the model on nuScenes' training slice with a batch size of 64 on 6 NVIDIA V100 GPUs. We use the AdamW optimizer with a learning rate of $2.5 \times 10^{-4}$ for 140 epochs, dropping the learning rate by 0.1 at epochs 90 and 120. Ground truth 3D bounding boxes are used to construct the ROI frustum of the objects in the frustum-based association step, ensuring that the radar cluster features are learned properly regardless of the image-based object detector performance. A random horizontal flip augmentation with a probability of 0.5 is used during training. We randomly flip both the input image and radar point cloud horizontally to improve the data diversity and the model's robustness to viewpoint variations.

\subsection{Testing Details}

We tested and evaluated the model on both nuScenes' validation and test slice. In contrast to the training process, no ground truth information is used during testing as we instead use the image-based object detector to generate the objects' ROI frustum. We also use horizontal flip augmentation, processing the original input as well as its horizontally flipped version and taking the average results. To evaluate the model performance, we use the nuScenes' official object detection evaluation metrics consisting of the mean average precision (mAP) and five true positive (TP) metrics: the mean average translation (mATE), scale (mASE), orientation (mAOE), velocity (mAVE), and attribute estimation error (mAAE). The overall model performance is measured by the nuScenes detection score (NDS), a weighted sum of the mAP and the five TP metrics described in Equation \ref{eq:nds} below
\begin{equation}
\begin{split}
    \text{NDS} &= 0.5 \text{mAP} + 0.1 \text{mATE} + 0.1 \text{mASE} \\
    &+ 0.1 \text{mAOE} + 0.1 \text{mAVE} + 0.1 \text{mAAE}.
\end{split}
\label{eq:nds}
\end{equation}

\section{Experiments}
\label{sec:experiments}

\begin{table*}[htbp]
    \caption{Performance comparison of different radar feature extraction strategies on nuScenes' official validation slice}
    \begin{center}
        \begin{tabular}{c|c|c|c|c|c|c|c}
            \hline
            \multirow{2}{*}{\textbf{Model}} & \textbf{Radar Feat.} & \multicolumn{2}{c}{\textbf{Performance $\uparrow$}} & \multicolumn{4}{|c}{\textbf{Error $\downarrow$}} \\
            \cline{3-8}
            & \textbf{Dimension} & \textbf{NDS} & \textbf{mAP} & \textbf{mATE} & \textbf{mAOE} & \textbf{mAVE} & \textbf{mAAE} \\
            \hline
            CenterFusion (baseline) \cite{nabati2021centerfusion} & 3 & 45.3 & 33.1 & \underline{0.646} & 0.556 & 0.513 & 0.143 \\
            Handcrafted & 13 & \textbf{46.5} & \textbf{33.5} & \textbf{0.642} & \textbf{0.519} & \underline{0.466} & \textbf{0.134} \\
            Learning-based & 1024 & 45.8 & 32.7 & 0.668 & 0.529 & \textbf{0.461} & \textbf{0.134} \\
            Handcrafted + Learning-based & 1037 & \underline{46.3} & \underline{33.2} & \underline{0.646} & \underline{0.520} & \underline{0.466} & \underline{0.135} \\
            \hline
            \multicolumn{8}{l}{*The best results are written in boldface and the second-best results are underlined.}
        \end{tabular}
        \label{tab:radar_feature_comp}
    \end{center}
\end{table*}

\begin{table*}[htbp]
    \caption{Ablation experiments results on nuScenes' official validation slice}
    \begin{center}
        \begin{tabular}{c|c|c|c|c|c|c|c}
            \hline
            \multirow{2}{*}{\textbf{Model}} & \multicolumn{2}{c}{\textbf{Performance $\uparrow$}} & \multicolumn{5}{|c}{\textbf{Error $\downarrow$}} \\
            \cline{2-8}
            & \textbf{NDS} & \textbf{mAP} & \textbf{mATE} & \textbf{mASE} & \textbf{mAOE} & \textbf{mAVE} & \textbf{mAAE} \\
            \hline
            CenterFusion \cite{nabati2021centerfusion} & 44.7 & 32.5 & 0.656 & 0.262 & 0.547 & 0.544 & 0.144 \\
            + 3$\times$3 Conv Layers in Secondary Regression Heads & 45.3 & 33.1 & 0.646 & \underline{0.262} & 0.556 & 0.513 & \underline{0.143} \\
            + Handcrafted Radar Cluster Features & 46.5 & 33.5 & \underline{0.642} & \textbf{0.261} & 0.519 & 0.466 & \textbf{0.134} \\
            + Modified Image-based Detector & \underline{47.9} & \underline{33.7} & 0.646 & \underline{0.262} & \underline{0.396} & \underline{0.444} & 0.152 \\
            + Uncertainty-aware 3D Confidence - ClusterFusion & \textbf{49.0} & \textbf{34.7} & \textbf{0.621} & \textbf{0.261} & \textbf{0.384} & \textbf{0.416} & 0.150 \\
            \hline
            \multicolumn{8}{l}{*The best results are written in boldface and the second-best results are underlined.}
        \end{tabular}
        \label{tab:ablation_study}
    \end{center}
\end{table*}

In this section, we investigated the performance of the three radar feature extraction strategies described in Section \ref{sec:method_radar_cluster_feature_extraction} to find the best-performing strategy. The best-performing strategy is then used as the radar feature extraction module of the final model, ClusterFusion. We tested ClusterFusion on the nuScenes' test slice and compared the results to the state-of-the-art methods on the nuScenes 3D detection task leaderboard. For fairness and rigor, we benchmarked ClusterFusion's performance twice: first against monocular camera and radar-monocular camera fusion-based methods, and second against methods that are not monocular camera-based. We also provided an analysis of ClusterFusion's runtime and compared it to the baseline model. To better understand the model's performance qualitatively, we provided some sample results produced by our model along with an analysis of them. Finally, ablation experiments are performed to see how the design choices affect the performance of the different radar feature extraction strategies as well as how ClusterFusion behaves in low-visibility conditions.

\subsection{Radar Feature Extractor Comparison}
\label{subsec:experiments_radar_feat_comp}

We trained four models, each using a different radar feature extraction strategy: the three strategies proposed in Section \ref{sec:method_radar_cluster_feature_extraction} and CenterFusion's \cite{nabati2021centerfusion} radar feature extraction strategy as a baseline. To isolate the effects of the modifications done to the image-based detector and the loss functions, we use CenterFusion's image-based object detector and loss functions for all the compared models. We however keep the secondary regression head architecture of five 3$\times$3 convolutional layers followed by a 1$\times$1 convolutional layer to ensure that the secondary regression heads are able to handle the more complex radar features.

We used a pre-trained CenterNet 3D object detection model that was provided by \cite{zhou2020tracking} and trained on the nuScenes dataset for 140 epochs as our models' image-based object detector. The models were trained further for 60 epochs on nuScenes' training slice using the AdamW optimizer with a batch size of 64 on 6 NVIDIA V100 GPUs and a learning rate of $5 \times 10^{-5}$, dropping it by 0.1 on epoch 45 and 55. To augment the data, improve the data diversity, and improve the model's robustness against viewpoint variations, we apply random horizontal flip and shift augmentation with a probability of 0.5 and 0.1 respectively. As with the final model, ground truth bounding boxes are used during training but not during testing. The models are then tested and evaluated on nuScenes' validation slice with flip augmentation. The test results are shown in Table \ref{tab:radar_feature_comp}. Note that the mASE metric is not used as the modification to the radar features does not affect 3D dimension estimation in any way.

All three proposed radar feature extraction strategies achieved higher NDS than the baseline strategy. Surprisingly, the handcrafted feature extraction strategy outperforms all other strategies, achieving the best performance in all metrics but mAVE. The handcrafted features might perform better as it is simpler than other features, having only 13 channels as opposed to over 1000 channels, especially considering that the image features have only 64 channels. This makes it easier for the secondary regression heads to learn the joint features from both modalities and generate more accurate predictions. These results show that all the proposed strategies are able to extract the local spatial features as well as the point-wise features of the radar point clouds and that the utilization of the aforementioned features leads to an all-around performance improvement, even on a semantic metric like the mAAE.

As the best-performing strategy, the handcrafted feature extraction strategy is used as the radar feature extraction module for the final model. The final model with the chosen strategy, ClusterFusion, is trained and tested following the details described in Section \ref{sec:implementation_details}. To highlight the contribution of each part, we trained several models with accumulative changes to the architecture. The performance of each model on nuScenes' validation slice is shown in Table \ref{tab:ablation_study}. The test results show that each change made to the CenterFusion architecture improves the overall performance of the model and that the final model achieves the best overall performance among all the models. The usage of the handcrafted radar feature extraction strategy in particular improves the orientation, velocity, and attribute estimation accuracy significantly. The usage of the modified image-based 3D object detector further boosts the orientation estimation accuracy but causes a drop in the attribute estimation accuracy. Finally, the depth estimate uncertainty-aware 3D detection confidence improves the last model's performance in all metrics and completes ClusterFusion.

\subsection{State-of-the-art Comparison}
\label{subsec:experiments_sota_comp}

\begin{table*}[htbp]
    \caption{Performance comparison with monocular camera and radar-monocular camera fusion 3D object detectors on nuScenes' official test slice}
    \begin{center}
        \begin{tabular}{c|c|c|c|c|c|c|c|c|c}
            \hline
            \multirow{2}{*}{\textbf{Model}} & \multirow{2}{*}{\textbf{Input}} & \textbf{Image-based} & \multicolumn{2}{c}{\textbf{Performance $\uparrow$}} & \multicolumn{5}{|c}{\textbf{Error $\downarrow$}} \\
            \cline{4-10}
            & & \textbf{Det. Backbone} & \textbf{NDS} & \textbf{mAP} & \textbf{mATE} & \textbf{mASE} & \textbf{mAOE} & \textbf{mAVE} & \textbf{mAAE} \\
            \hline
            CenterNet \cite{zhou2019objects} & C & Hourglass \cite{newell2016stacked} & 40.0 & 33.8 & 0.658 & 0.255 & 0.629 & 1.629 & 0.142 \\
            FCOS3D \cite{wang2021fcos3d} & C & ResNet101 \cite{he2016deep} & 42.8 & 35.8 & 0.690 & \underline{0.249} & 0.452 & 1.434 & \underline{0.124} \\
            PGD \cite{wang2022probabilistic} & C & ResNet101 \cite{he2016deep} & 44.8 & \underline{38.6} & 0.626 & \textbf{0.245} & 0.451 & 1.509 & 0.127 \\
            DD3D \cite{park2021pseudo} & C & V2-99 \cite{lee2020centermask} & \underline{47.7} & \textbf{41.8} & \textbf{0.572} & \underline{0.249} & \textbf{0.368} & \underline{1.014} & \underline{0.124} \\
            ClusterFusion (ours) & C,R & DLA-34 \cite{yu2018deep} & \textbf{48.7} & 34.1 & \underline{0.587} & 0.257 & \underline{0.424} & \textbf{0.461} & \textbf{0.108} \\
            \hline
            CenterFusion \cite{nabati2021centerfusion} & C,R & DLA-34 \cite{yu2018deep} & 44.9 & 32.6 & 0.631 & 0.261 & \underline{0.516} & 0.614 & \underline{0.115} \\
            CenterTransFuser \cite{li2023centertransfuser} & C,R & DLA-34 \cite{yu2018deep} & 47.1 & 34.7 & 0.628 & \textbf{0.252} & 0.523 & \underline{0.527} & 0.135 \\
            RCBEV \cite{zhou2023bridging} & C,R & Swin-T \cite{liu2021swin} & \underline{48.6} & \textbf{40.6} & \textbf{0.484} & \underline{0.257} & 0.587 & 0.702 & 0.140 \\
            PGD + RADIANT \cite{long2023radiant} & C,R & ResNet101 \cite{he2016deep} & - & \underline{38.0} & 0.609 & - & - & - & - \\
            ClusterFusion (ours) & C,R & DLA-34 \cite{yu2018deep} & \textbf{48.7} & 34.1 & \underline{0.587} & \underline{0.257} & \textbf{0.424} & \textbf{0.461} & \textbf{0.108} \\
            \hline
            \multicolumn{8}{l}{*The best results are written in boldface and the second-best results are underlined.}
        \end{tabular}
        \label{tab:monocular_performance_test_slice}
    \end{center}
\end{table*}

\begin{table*}[htbp]
    \caption{Performance comparison with 3D object detectors that are not monocular camera-based on nuScenes' official test slice}
    \begin{center}
        \begin{tabular}{c|c|c|c|c|c|c|c|c|c|c|c}
            \hline
            \multirow{2}{*}{\textbf{Model}} & \multirow{2}{*}{\textbf{Input}} & \textbf{Image-based} & \textbf{Multi} & \textbf{Temporal} & \multicolumn{2}{c}{\textbf{Performance $\uparrow$}} & \multicolumn{5}{|c}{\textbf{Error $\downarrow$}} \\
            \cline{6-12}
            & & \textbf{Det. Backbone} & \textbf{Cams} & \textbf{Info.} & \textbf{NDS} & \textbf{mAP} & \textbf{mATE} & \textbf{mASE} & \textbf{mAOE} & \textbf{mAVE} & \textbf{mAAE} \\
            \hline
            CenterPoint \cite{yin2021center} & L & - & - & \checkmark & \textbf{67.3} & \textbf{60.3} & \textbf{0.262} & \textbf{0.239} & \textbf{0.361} & \textbf{0.288} & \underline{0.136} \\
            ClusterFusion (ours) & C,R & DLA-34 \cite{yu2018deep} & - & - & \underline{48.7} & \underline{34.1} & \underline{0.587} & \underline{0.257} & \underline{0.424} & \underline{0.461} & \textbf{0.108} \\
            \hline
            DETR3D \cite{wang2022detr3d} & C & V2-99 \cite{lee2020centermask} & \checkmark & - & 47.9 & 41.2 & 0.641 & 0.255 & 0.394 & 0.845 & 0.133 \\
            BEVDet \cite{huang2021bevdet} & C & V2-99 \cite{lee2020centermask} & \checkmark & - & \underline{48.8} & \underline{42.4} & \textbf{0.524} & \textbf{0.242} & \textbf{0.373} & 0.950 & 0.148 \\
            PETR \cite{liu2022petr} & C & V2-99 \cite{lee2020centermask} & \checkmark & - & \textbf{50.4} & \textbf{44.1} & 0.593 & \underline{0.249} & \underline{0.384} & \underline{0.808} & \underline{0.132} \\
            ClusterFusion (ours) & C,R & DLA-34 \cite{yu2018deep} & - & - & 48.7 & 34.1 & \underline{0.587} & 0.257 & 0.424 & \textbf{0.461} & \textbf{0.108} \\
            \hline
            BEVDet4D \cite{huang2022bevdet4d} & C & Swin-B \cite{lee2020centermask} & \checkmark & \checkmark & 56.9 & 45.1 & 0.511 & \textbf{0.241} & 0.386 & 0.301 & \underline{0.121} \\
            BEVFormer \cite{li2022bevformer} & C & V2-99 \cite{lee2020centermask} & \checkmark & \checkmark & 56.9 & 48.1 & 0.582 & 0.256 & 0.375 & 0.378 & 0.126 \\
            BEVDepth \cite{li2023bevdepth} & C & ConvNeXt-B \cite{liu2022convnet} & \checkmark & \checkmark & 60.9 & 52.0 & 0.445 & \underline{0.243} & 0.352 & 0.347 & 0.127 \\
            SOLOFusion \cite{park2023time} & C & ConvNeXt-B \cite{liu2022convnet} & \checkmark & \checkmark & 61.9 & 54.0 & 0.453 & 0.257 & 0.376 & 0.276 & 0.148 \\
            BEVFormerV2Opt \cite{yang2023bevformer} & C & InternImage-XL \cite{wang2023internimage} & \checkmark & \checkmark & 64.8 & 58.0 & 0.448 & 0.262 & \underline{0.342} & \underline{0.238} & 0.128 \\
            BEVDet-Gamma \cite{huang2022bevpoolv2} & C & Swin-B \cite{liu2021swin} & \checkmark & \checkmark & \underline{66.4} & \underline{58.6} & \textbf{0.375} & \underline{0.243} & 0.377 & \textbf{0.174} & 0.123 \\
            VideoBEV \cite{han2023exploring} & C & ConvNeXt-B \cite{liu2022convnet} & \checkmark & \checkmark & \textbf{67.0} & \textbf{59.2} & \underline{0.385} & 0.246 & \textbf{0.323} & \textbf{0.174} & 0.137 \\
            ClusterFusion (ours) & C,R & DLA-34 \cite{yu2018deep} & - & - & 48.7 & 34.1 & 0.587 & 0.257 & 0.424 & 0.461 & \textbf{0.108} \\
            \hline
            MVFusion \cite{wu2023mvfusion} & C,R & V2-99 \cite{lee2020centermask} & \checkmark & - & 51.7 & 45.3 & 0.569 & 0.246 & \textbf{0.379} & 0.781 & 0.128 \\
            TransCAR \cite{pang2023transcar} & C,R & ResNet101 \cite{he2016deep} & \checkmark & - & 52.2 & 42.2 & 0.630 & 0.260 & 0.383 & 0.495 & 0.121 \\
            CRAFT \cite{kim2023craft} & C,R & DLA-34 \cite{yu2018deep} & \checkmark & - & 52.3 & 41.1 & 0.467 & 0.268 & 0.456 & 0.519 & 0.114 \\
            X3KD \cite{klingner2023x3kd} & C,R & ResNet101 \cite{he2016deep} & \checkmark & - & 55.3 & 44.1 & 0.499 & 0.257 & 0.435 & 0.378 & \textbf{0.107} \\
            RC-BEVFusion \cite{stacker2023rc} & C,R & Swin-T \cite{liu2021swin} & \checkmark & - & 56.7 & 47.6 & 0.444 & \underline{0.244} & 0.462 & 0.439 & 0.128 \\
            RCM-Fusion \cite{kim2023rcm} & C,R & ResNet101 \cite{he2016deep} & \checkmark & - & 58.0 & 49.3 & 0.485 & 0.255 & 0.386 & 0.421 & 0.115 \\
            CRN \cite{kim2023crn} & C,R & ConvNeXt-B \cite{liu2022convnet} & \checkmark & - & \underline{62.4} & \underline{57.5} & \underline{0.416} & 0.264 & 0.456 & \underline{0.365} & 0.130 \\
            HVDetFusion \cite{lei2023hvdetfusion} & C,R & InternImage-B \cite{wang2023internimage} & \checkmark & \checkmark & \textbf{67.4} & \textbf{60.4} & \textbf{0.379} & \textbf{0.243} & \underline{0.382} & \textbf{0.172} & 0.132 \\
            ClusterFusion (ours) & C,R & DLA-34 \cite{yu2018deep} & - & - & 48.7 & 34.1 & 0.587 & 0.257 & 0.424 & 0.461 & \underline{0.108} \\
            \hline
            \multicolumn{8}{l}{*The best results are written in boldface and the second-best results are underlined.}
        \end{tabular}
        \label{tab:all_performance_test_slice}
    \end{center}
\end{table*}

The final model, ClusterFusion, uses the handcrafted feature extraction strategy as its radar feature extraction module and is tested on nuScenes' official test slice and compared with other methods on the nuScenes 3D object detection task leaderboard. Table \ref{tab:monocular_performance_test_slice} shows the performance of ClusterFusion as well as the state-of-the-art monocular camera and radar-monocular camera fusion 3D object detectors on nuScenes' test slice.

ClusterFusion achieved the state-of-the-art with the best overall performance among radar-monocular camera fusion methods, leading with 48.7\% NDS, 0.424 mAOE, 0.461 mAVE, and 0.108 mAAE. It is however inferior to RCBEV \cite{zhou2023bridging} in terms of mATE and mAP in which it scored 0.587 and 0.257 respectively, both of which can be attributed to ClusterFusion having inferior object localization ability. RCBEV not only uses a stronger image backbone of Swin Transformer-Tiny \cite{liu2021swin} but also uses BEV image features and performs fusion on the BEV plane, allowing for better object localization. When compared against CenterFusion \cite{nabati2021centerfusion} and CenterTransFuser \cite{li2023centertransfuser} that both use the same backbone of DLA-34 \cite{yu2018deep} and perform fusion on the image plane, ClusterFusion shows better overall performance signified by its better NDS, mATE, mAOE, mAVE, and mAAE. The most significant improvements are observed in orientation, velocity, and attribute estimation which are all the metrics predicted by the secondary regression heads that are influenced directly by the radar features. This shows that extracting the local spatial features and the point-wise features directly from the radar point cloud clusters is beneficial to the detection performance, particularly to the orientation, velocity, and attribute estimation accuracy. This also suggests that the proposed radar feature extraction strategy used by ClusterFusion can effectively leverage the information contained in the radar point cloud clusters.

Similarly, when compared to monocular camera methods, ClusterFusion achieved the best overall performance with the best NDS, mAVE, and mAAE. However, it fell short of DD3D \cite{park2021pseudo} in mAP, mATE, and mAOE. Again, this can be attributed to DD3D's superior ability in localizing objects due to its pretraining using lidar data as depth ground truths as well as its stronger backbone of VoVNet V2-99 \cite{lee2020centermask}. When compared against CenterNet \cite{zhou2019objects} which is the base of ClusterFusion's image-based detector, ClusterFusion performs better in all metrics, with the most significant improvements observed in mAOE, mAVE, and mAAE which are the metrics predicted directly by the radar-influenced secondary regression heads.

Table \ref{tab:all_performance_test_slice} shows the performance comparison between ClusterFusion and other 3D object detectors that are not based on monocular cameras. Compared to CenterPoint \cite{yin2021center}, a lidar-based method, ClusterFusion is inferior in every metric save for mAAE. The highly accurate lidar point cloud enables CenterPoint to produce high-accuracy detection, localization, dimension estimation, and even velocity estimation by incorporating temporal information from multiple frames of data. 

When compared to recent multiple camera-based methods such as DETR3D \cite{wang2022detr3d}, BEVDet \cite{huang2021bevdet}, and PETR \cite{liu2022petr}, ClusterFusion achieved inferior performance in almost all metrics but mAVE and mAAE. Not only do these methods use features from multiple images, these methods also use the stronger VoVNet V2-99 \cite{lee2020centermask} backbone. BEVDet additionally uses BEV image features that enable better localization performance as shown by its lowest mATE and mAOE among other methods. Nevertheless, ClusterFusion proves superior in terms of mAVE and mAAE due to its use of radar information that includes radial velocity measurements.

However, multiple camera-based methods that use temporal information from multiple frames of input \cite{huang2022bevdet4d, li2022bevformer, li2023bevdepth, park2023time, yang2023bevformer, huang2022bevpoolv2, han2023exploring} outperform ClusterFusion even in terms of mAVE. Moreover, all of these methods use BEV image features, further enlarging the performance gap between them and ClusterFusion with their better localization performance. However, ClusterFusion still achieved better mAAE over them. It might be important to note that the image-based detector backbone used by ClusterFusion, DLA-34 \cite{yu2018deep}, is also weaker than the backbones used by the multiple camera-based methods shown in Table \ref{tab:all_performance_test_slice} that use newer and stronger backbones such as VoVNet V2-99 \cite{lee2020centermask}, Swin Transformer-Base \cite{liu2021swin}, ConvNeXt-Base \cite{liu2022convnet}, and InternImage-XL \cite{wang2023internimage}.

Among the compared radar-multiple camera fusion 3D object detection methods \cite{wu2023mvfusion, pang2023transcar, kim2023craft, klingner2023x3kd, stacker2023rc, kim2023rcm, kim2023crn, lei2023hvdetfusion}, it might be important to note that MVFusion \cite{wu2023mvfusion}, TransCAR \cite{pang2023transcar}, and CRAFT \cite{kim2023craft} do not use BEV image features while all the other methods do use BEV image features. The three non-BEV methods all achieved inferior overall performance to the BEV methods. The state-of-the-art radar-multiple camera fusion 3D object detection method, HVDetFusion \cite{lei2023hvdetfusion}, additionally uses temporal information from multiple frames of images. ClusterFusion uses a single camera without the use of BEV features or temporal information. Compared to the radar-multiple camera fusion 3D object detection methods presented in Table \ref{tab:all_performance_test_slice}, it achieved generally inferior performance on almost all the metrics but mAAE where it achieved the second best result to X3KD \cite{klingner2023x3kd}. X3KD might benefit from the fact that it distills knowledge from a lidar-based detector and a camera-based instance segmentation model during training, both of which require additional data and labels. Even with all the disadvantages, ClusterFusion achieved a competitive or even better mAAE when compared to other methods that are not monocular camera-based.

\subsection{Runtime Analysis}

ClusterFusion achieved the results discussed in Section \ref{subsec:experiments_sota_comp} with minimal additional computations over CenterFusion. The inference runtime comparison between ClusterFusion and the baseline model CenterFusion \cite{nabati2021centerfusion} is shown in Table \ref{tab:runtime_comparison}. The comparison is done on a PC running an Intel Core i9-10920X CPU with 125 GB RAM and an NVIDIA GeForce RTX 3090. We compared the runtime of the original CenterFusion implementation by \cite{nabati2021centerfusion}, our implementation of CenterFusion, and ClusterFusion which is built upon our implementation of CenterFusion.

\begin{table}[htbp]
    \caption{Inference runtime comparison with the baseline model}
    \begin{center}
        \begin{tabular}{c|c|c}
            \hline
            \textbf{Model} & \textbf{$\mathbf{t}$ (ms)} & \textbf{FPS} \\
            \hline
            CenterFusion \cite{nabati2021centerfusion} & 310.8 & 3.22 \\
            CenterFusion (our implementation) & 93.7 & 10.67 \\
            ClusterFusion (ours) & 101.4 & 9.86 \\
            \hline
        \end{tabular}
        \label{tab:runtime_comparison}
    \end{center}
\end{table}

We found that our implementation of CenterFusion is over 3 times faster than the original implementation, thanks to our extensive use of vectorization especially in the frustum-based association and heatmap conversion mechanisms. ClusterFusion, while slower than our implementation of CenterFusion, adds only 7.7 ms of computation per frame or equivalent to an 8.21\% increase of computation time and is only 0.86 FPS slower. It trades the additional computation time for a 3.8\% increase in NDS and a 1.5\% increase in mAP.

\subsection{Qualitative Analysis}

\begin{figure*}[!htbp]
    \centering
    \includegraphics[width=\textwidth]{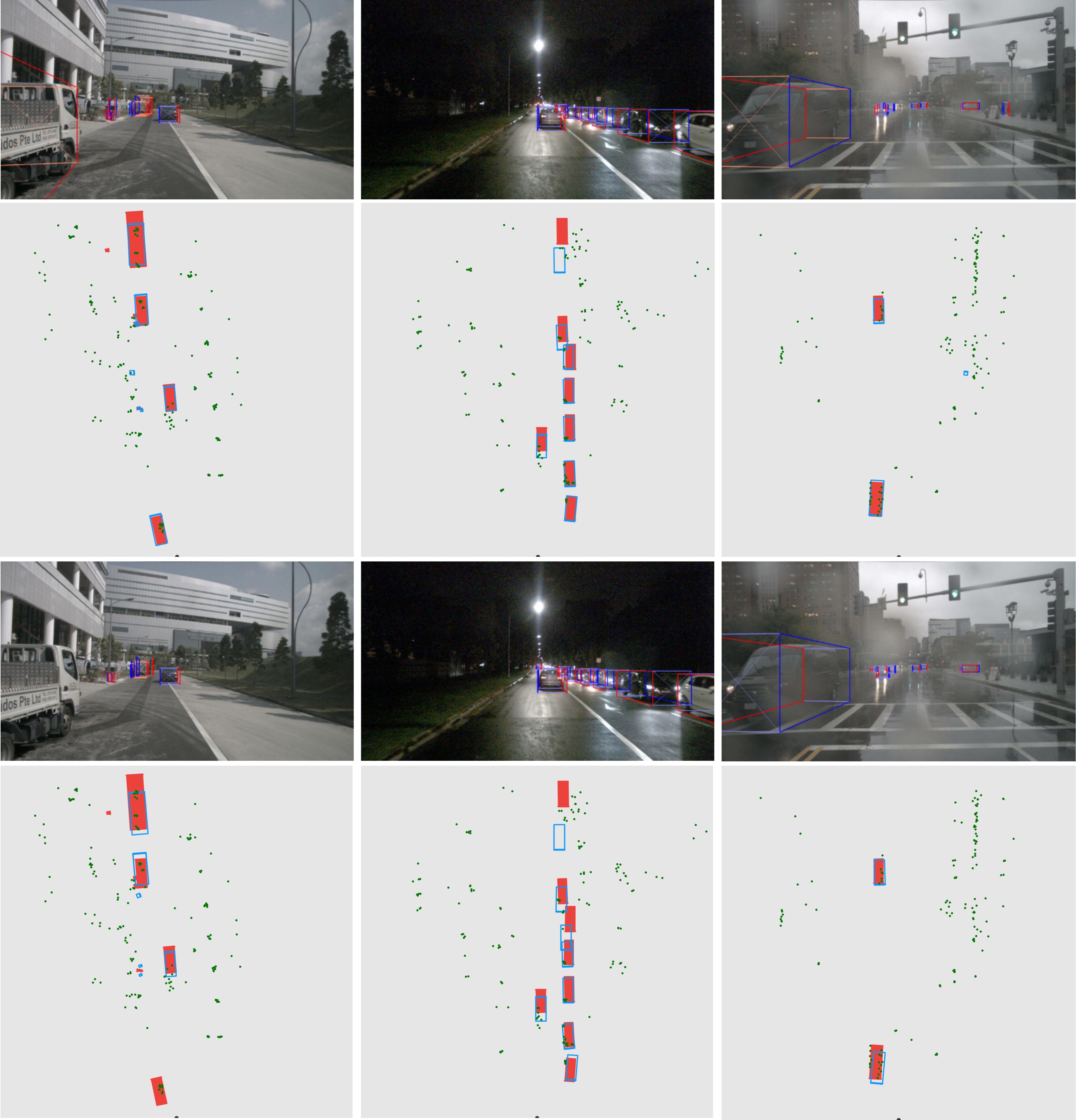}
    \caption{Sample detection results from the nuScenes dataset, obtained using our proposed method (top) and the baseline method CenterFusion (bottom). Our proposed method performs better than the baseline method, especially in detecting obstructed, truncated, and faraway objects.}
    \label{fig:qualitative_results}
\end{figure*}

Figure \ref{fig:qualitative_results} shows a comparison of the detection results obtained using ClusterFusion and CenterFusion \cite{nabati2021centerfusion}, the baseline architecture, during daytime, nighttime, and in rain. Our proposed method can perform well even in low-visibility conditions. Our proposed method performs better than CenterFusion in detecting obstructed and truncated objects. It is also better for estimating objects' position and orientation even when they are far away from the ego vehicle.

\subsection{Ablation Study}
\label{sec:ablation_study}

\begin{table*}[htbp]
    \caption{Performance comparison of handcrafted feature combinations}
    \begin{center}
        \begin{tabular}{c|c|c|c|c|c|c|c|c|c|c|c|c|c}
            \hline
            \multirow{2}{*}{\textbf{Model}} & \multicolumn{6}{c|}{\textbf{Features}} & \multirow{2}{*}{\textbf{Dim}} & \multicolumn{2}{|c}{\textbf{Performance $\uparrow$}} & \multicolumn{4}{|c}{\textbf{Error $\downarrow$}} \\
            \cline{2-7}
            \cline{9-14}
            & \textbf{Mean} & \textbf{Min} & \textbf{Max} & \textbf{Med} & \textbf{Var} & \textbf{Ort} &
            & \textbf{NDS} & \textbf{mAP} & \textbf{mATE} & \textbf{mAOE} & \textbf{mAVE} & \textbf{mAAE} \\
            \hline
            Mean & \checkmark & \checkmark & \checkmark & - & - & - & 12 & \textbf{38.5} & \textbf{31.4} & \textbf{0.702} & \underline{0.513} & \underline{1.232} & 0.245 \\
            MeanOrt & \checkmark & \checkmark & \checkmark & - & - & \checkmark & 13 & 37.3 & 29.5 & 0.760 & \textbf{0.491} & \textbf{1.202} & \textbf{0.236} \\
            MedianOrt & - & \checkmark & \checkmark & \checkmark & - & \checkmark & 13 & 36.7 & 30.0 & 0.761 & 0.561 & 1.272 & \underline{0,239} \\
            Complete & \checkmark & \checkmark & \checkmark & \checkmark & \checkmark & \checkmark & 21 & \underline{37.5} & \underline{30.7} & \underline{0.730} & 0.537 & 1.337 & 0.256 \\
            \hline
            \multicolumn{14}{l}{*The best results are written in boldface and the second-best results are underlined.}
        \end{tabular}
        \label{tab:radar_handcrafted_alts}
    \end{center}
\end{table*}

\begin{table*}[htbp]
    \caption{Performance comparison of learning-based radar feature extractor architecture configurations}
    \begin{center}
        \begin{tabular}{c|c|c|c|c|c|c|c|c|c|c}
            \hline
            \multirow{3}{*}{\textbf{Model}} & \multicolumn{4}{c|}{\textbf{Configuration}} & \multicolumn{2}{|c}{\textbf{Performance $\uparrow$}} & \multicolumn{4}{|c}{\textbf{Error $\downarrow$}} \\
            \cline{2-5}
            \cline{6-11}
            & \textbf{Kernel} & \textbf{\# KPConv} & \multicolumn{2}{c|}{\textbf{Feature Map Dim}} & \multirow{2}{*}{\textbf{NDS}} & \multirow{2}{*}{\textbf{mAP}} & \multirow{2}{*}{\textbf{mATE}} & \multirow{2}{*}{\textbf{mAOE}} & \multirow{2}{*}{\textbf{mAVE}} & \multirow{2}{*}{\textbf{mAAE}} \\
            \cline{4-5}
            & \textbf{Size} & \textbf{Layers} & \textbf{First} & \textbf{Output} & & & & & & \\
            \hline
            Lite & 8 & 4 & 8 & 64 & \underline{38.2} & 29.7 & \underline{0.718} & \textbf{0.445} & 1.580 & \underline{0.235} \\
            Medium & 15 & 5 & 32 & 512 & 37.5 & \underline{30.5} & 0.749 & 0.535 & \underline{1.345} & 0.231 \\
            Large & 15 & 5 & 64 & 1024 & \textbf{38.9} & \textbf{31.0} & \textbf{0.712} & \underline{0.472} & \textbf{1.314} & \textbf{0.208} \\
            \hline
            \multicolumn{11}{l}{*The best results are written in boldface and the second-best results are underlined.}
        \end{tabular}
        \label{tab:radar_learned_alts}
    \end{center}
\end{table*}

\begin{table*}[htbp]
    \caption{Results of ablation experiments under different conditions}
    \begin{center}
        \begin{tabular}{c|c|c|c|c|c|c|c|c}
            \hline
            \multirow{2}{*}{\textbf{Model}} & \textbf{Number of} & \multicolumn{2}{c}{\textbf{Performance $\uparrow$}} & \multicolumn{5}{|c}{\textbf{Error $\downarrow$}} \\
            \cline{3-9}
            & \textbf{Scenes} & \textbf{NDS} & \textbf{mAP} & \textbf{mATE} & \textbf{mASE} & \textbf{mAOE} & \textbf{mAVE} & \textbf{mAAE} \\
            \hline
            Ideal & 112 & \textbf{49.2} & \underline{35.0} & \textbf{0.615} & \textbf{0.258} & \textbf{0.386} & \textbf{0.406} & \underline{0.159} \\
            Rain & 24 & \underline{49.0} & \textbf{35.2} & \underline{0.644} & \underline{0.283} & \underline{0.390} & \underline{0.432} & \textbf{0.110} \\
            Night & 15 & 26.1 & 16.3 & 0.741 & 0.477 & 0.647 & 0.863 & 0.478 \\
            \hline
            \multicolumn{8}{l}{*The best results are written in boldface and the second-best results are underlined.}
        \end{tabular}
        \label{tab:ablation_study_night_rain}
    \end{center}
\end{table*}

\subsubsection{Radar Feature Configurations}

In our search for the best configuration for each radar feature extraction strategy, we trained and tested several models with different configurations for each strategy and compared their performance against each other. To save time, we use only a tenth of the nuScenes' training and validation slice to train and test the models. We use training and test configurations similar to the configuration described in Section \ref{subsec:experiments_radar_feat_comp}, but we trained the models for 100 epochs instead.

For the handcrafted feature extraction strategy, we compared several feature combinations. Table \ref{tab:radar_handcrafted_alts} shows the feature combinations along with their respective detection performance. The \texttt{Mean} model that only uses the mean, minimum, and maximum values of the point-wise radar features achieved the best NDS and mAP as it boasts a superior position estimation accuracy. The \texttt{MeanOrt} model that extends the \texttt{Mean} model by adding the orientation estimate feature achieved the best objects' orientation, velocity, and attribute estimation performance but achieved a lower NDS and mAP. This shows that the orientation estimate feature can effectively represent the radar point cloud cluster's orientation and improve detection performance. We tested the \texttt{MedianOrt} model, a variation of the \texttt{MeanOrt} model that uses median instead of mean, to see if the median can represent the radar clusters' feature distribution better. However, it turned out that this model performed the worst among all the models. Lastly, we tested the \texttt{Complete} model that uses the variance value on top of using both mean and median but it was outperformed by the \texttt{Mean} and the \texttt{MeanOrt} model. We chose the \texttt{MeanOrt} model's feature combination as the best configuration for the handcrafted radar feature extraction strategy as it gave the best orientation, velocity, and attribute estimation performance.

For the learning-based radar feature extraction strategy, we compared several architecture configurations. Table \ref{tab:radar_learned_alts} shows the architecture configurations along with their detection performance. The result is quite straightforward, as the largest model with the largest kernel size, the most number of KPConv layers, and the largest feature map dimensions achieved the best performance among all the models. As such, we use the architecture configurations used by the \texttt{Large} model as the best configuration for the learning-based radar feature extraction strategy.

\subsubsection{Performance in Low-Visibility Conditions}

To study ClusterFusion's performance in low-visibility conditions such as in the rain or during nighttime, we tested the model on three different subsets of the nuScenes' validation slice. The \texttt{Rain} subset contains only rainy scenes, the \texttt{Night} subset contains only nighttime scenes, and the \texttt{Ideal} subset contains all the other scenes. Table \ref{tab:ablation_study_night_rain} shows the performance of the model on the three datasets.

The model's performance does not really degrade in the rain, it only showed slight all-around performance degradation in all the metrics but mAP and mAAE where it instead showed slight improvement. However, the model's performance degrades significantly during nighttime, achieving only about half the NDS and mAP of its performance in ideal conditions. This performance degradation might suggest that ClusterFusion relies heavily upon the image-based detector, as when it can not produce adequate preliminary 3D detections it can not associate radar clusters properly. Thus, the information from radar can not be utilized well. However, the poor performance during nighttime might also be explained by the small size of the dataset, skewing the performance distribution.

\section{Conclusions}
\label{sec:conclusions}

We proposed ClusterFusion, a radar-monocular camera fusion-based 3D object detection method that achieved state-of-the-art performance with the best NDS, mAOE, mAVE, and mAAE as well as the second-best mATE and mASE among all radar-monocular camera methods on nuScenes' object detection leaderboard. Compared to the state-of-the-art 3D object detection methods that are not based on monocular cameras, ClusterFusion achieved a competitive attribute estimation performance. ClusterFusion performs cross-modal feature fusion on the image plane, but in contrast to previous works, it performs feature extraction on radar clusters directly in their point cloud form rather than in their image-like projected form, preserving the local spatial features of the clusters that are otherwise lost in the projection process. We also investigated and compared the performance of three radar feature extraction strategies: a handcrafted strategy, a learning-based strategy leveraging KPConv \cite{thomas2019kpconv}, and a combination of both, and found that the handcrafted strategy yields the best result.

The main shortcoming of ClusterFusion is that it relies on image-based preliminary 3D object detections to filter and cluster the radar point cloud. Consequently, when the accuracy of the image-based preliminary detections is subpar, ClusterFusion struggles to utilize radar information effectively. Incorporating radar information in the generation of preliminary 3D object detections might help alleviate this issue. Furthermore, ClusterFusion fuses the image features and radar features on the image plane using only a simple concatenation. The use of a more effective cross-modal feature fusion strategy can be a topic for further investigation. The use of BEV image features could prove both promising and straightforward as the radar feature extraction framework can theoretically be adapted directly for use on the BEV plane. Additionally, the use of temporal information along with a stronger backbone might help improve the image-based detector's performance.

\bibliography{IEEEabrv, references}

\EOD

\end{document}